\documentclass{article}

\usepackage{microtype}
\usepackage{graphicx}
\usepackage{subcaption}
\usepackage{booktabs} %

\usepackage{hyperref}
\usepackage{url}

\usepackage{amsmath,amssymb,amsthm}
\usepackage{graphicx}
\usepackage{booktabs}
\usepackage{adjustbox}
\usepackage{multirow}
\usepackage{makecell}
\usepackage{wrapfig}
\usepackage{enumitem}
\usepackage{subcaption}
\usepackage[table]{xcolor}
\definecolor{lightgreen}{RGB}{217,242,217}

\DeclareMathOperator{\Var}{Var}

\usepackage[preprint]{icml2026}

\usepackage{amsmath}
\usepackage{amssymb}
\usepackage{mathtools}
\usepackage{amsthm}

\usepackage[capitalize,noabbrev]{cleveref}

\theoremstyle{plain}

\theoremstyle{definition}

\theoremstyle{remark}

\usepackage[textsize=tiny]{todonotes}

\icmltitlerunning{A Geometry-Based View of Mahalanobis OOD Detection}

\begin{document}

\twocolumn[
  \icmltitle{A Geometry-Based View of Mahalanobis OOD Detection}

  \icmlsetsymbol{equal}{*}

  \begin{icmlauthorlist}
    \icmlauthor{Denis Janiak}{pwr}
    \icmlauthor{Jakub Binkowski}{pwr}
    \icmlauthor{Tomasz Kajdanowicz}{pwr}
  \end{icmlauthorlist}

  \icmlaffiliation{pwr}{Department of Artificial Intelligence, Wrocław University of Science and Technology, Wrocław, Poland}

  \icmlcorrespondingauthor{Denis Janiak}{denis.janiak@pwr.edu.pl}

  \icmlkeywords{Machine Learning, Out-of-Distribution Detection, Mahalanobis Distance}

  \vskip 0.3in
]

\printAffiliationsAndNotice{}  %

\begin{abstract}
Out-of-distribution (OOD) detection is critical for reliable deployment of vision models. Mahalanobis-based detectors remain strong baselines, yet their performance varies widely across modern pretrained representations, and it is unclear which properties of a feature space cause these methods to succeed or fail. We conduct a large-scale study across diverse foundation-model backbones and Mahalanobis variants. First, we show that Mahalanobis-style OOD detection is not universally reliable: performance is highly representation-dependent and can shift substantially with pretraining data and fine-tuning regimes. Second, we link this variability to in-distribution geometry and identify a two-term ID summary that consistently tracks Mahalanobis OOD behavior across detectors: within-class spectral structure and local intrinsic dimensionality. Finally, we treat normalization as a geometric control mechanism and introduce radially scaled $\ell_2$ normalization, $\phi_\beta(z)=z/\|z\|^\beta$, which preserves directions while contracting or expanding feature radii. Varying $\beta$ changes the radii while preserving directions, so the same quadratic detector sees a different ID geometry. We choose $\beta$ from ID-only geometry signals and typically outperform fixed normalization baselines.
\end{abstract}

\section{Introduction}

\begin{figure*}[t]
    \centering
    \includegraphics[width=0.8\linewidth]{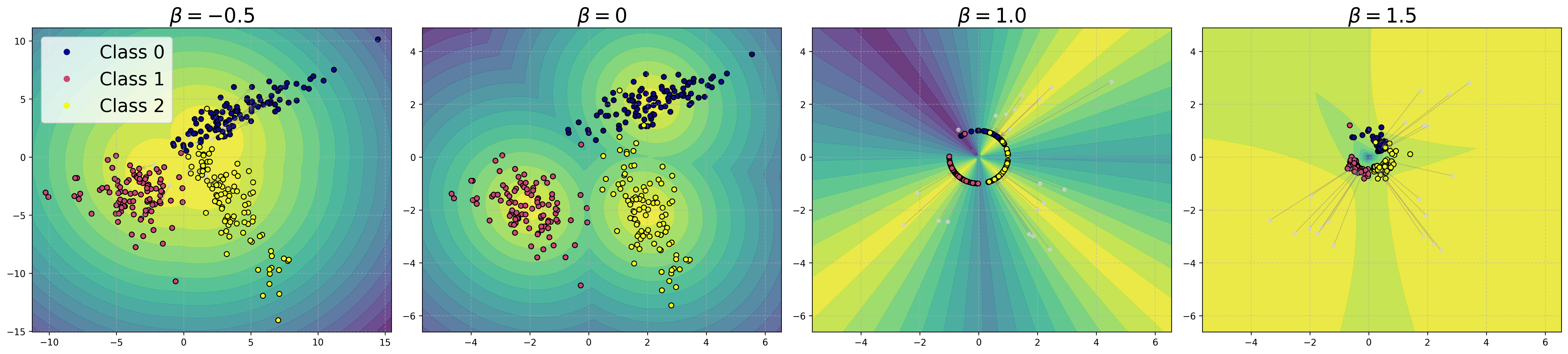}
    \caption{
    \textbf{Effect of radially scaled $\ell_2$ normalization on feature geometry and Mahalanobis boundaries.}
    Visualization in 2D of the induced geometry and the resulting Mahalanobis decision regions. Gray arrows indicate the radial mapping from the original to the transformed space. Larger $\beta$ contracts feature norms and tightens clusters, producing more localized decision regions; smaller $\beta$ expands norms and broadens them. Choosing $\beta$ appropriately can reduce ID--OOD overlap and improve OOD detection.
    }
    \label{fig:decision_boundary}
    \vspace{-0.5em}
\end{figure*}

Mahalanobis scores are among the simplest post-hoc detectors for out-of-distribution (OOD) detection, yet they remain surprisingly competitive on modern vision backbones \citep{lee_simple_2018, mueller_mahalanobis_2025, mueller_how_2024}. At the same time, their behavior is highly representation-dependent: the same quadratic detector can perform well on one pretrained model and fail on another, and performance can shift sharply with changes in pretraining data or fine-tuning regime. This sensitivity makes Mahalanobis-based OOD detection difficult to deploy reliably and raises a basic question: \emph{which properties of an in-distribution feature space determine when a Mahalanobis detector succeeds or fails?}

We study Mahalanobis OOD detection through the lens of representation geometry. Across self supervised and foundation model representations, we show that geometric structure accounts for much of the observed cross model variation. In particular, a compact ID-only summary combining local intrinsic dimensionality and within-class spectral decay strongly predicts Mahalanobis OOD performance across variants. This connects detector reliability to measurable properties of the in-distribution feature space.

Motivated by this geometric view, we introduce a simple post-hoc control mechanism that changes the geometry presented to the same quadratic detector. We use radially scaled $\ell_2$ normalization,
$\phi_\beta(z)=z/\|z\|^\beta$,
which preserves feature directions while contracting or expanding radii. Unlike prior work that modifies the scoring rule \citep{ren_simple_2021} or fixes normalization to the unit sphere \citep{mueller_mahalanobis_2025}, varying $\beta$ provides a continuous way to reshape radial geometry without altering the detector form. Figure~\ref{fig:decision_boundary} illustrates how $\beta$ tightens or spreads decision regions by changing feature radii. Empirically, adjusting $\beta$ induces structured, model-specific changes in both geometry and OOD performance. Leveraging the same ID-only geometry signals, we propose a practical procedure to select $\beta$ without access to OOD samples, often improving over fixed baselines such as $\beta=0$ (standard features) and $\beta=1$ (unit-sphere normalization).

Our main contributions are:
\begin{enumerate}[nosep,itemsep=0.5em]
    \item A broad benchmark of Mahalanobis-style OOD detectors across diverse SSL/foundation models, including a per-dimension analysis of detector behavior.
    \item An empirical link between Mahalanobis OOD performance and ID geometry, including an ID-only summary that consistently predicts performance across detector variants.
    \item A geometric control mechanism via $\beta$-scaled radial normalization, together with an ID-only $\beta$ selection rule that approaches oracle tuning without requiring OOD access.
\end{enumerate}

\section{Related Work}
OOD detection is essential for ensuring the reliability of machine learning systems in real-world deployment~\citep{10.5555/3540261.3540802}. Its goal is to identify whether inputs stem from the training distribution, thus preventing overconfident predictions on unexpected data~\citep{yang2024generalized}. Post-hoc, training-free methods are particularly effective, as they combine efficiency with robustness without altering the model~\citep{xu_scaling_2023}. Among OOD detection methods, Mahalanobis distance has become a cornerstone~\citep{lee_simple_2018}, with several refinements improving its robustness and performance. The standard Mahalanobis distance (MD) uses class-conditional covariance estimates to measure the distance of a sample from each class mean. In contrast, the Relative Mahalanobis distance (RMD)~\citep{ren_simple_2021} compares each class-specific distance to a single global Gaussian fitted to all in-distribution (ID) data, effectively normalizing class distances against a global reference. Mahalanobis++~\citep{mueller_mahalanobis_2025} further improves performance by L2-normalizing features, making them adhere more closely to the Gaussian assumptions underlying the Mahalanobis distance.
However, our study reveals broader insight into the influence of normalization when computing Mahalanobis distance, particularly in the context of vision models.

Vision OOD detection has shifted toward leveraging large-scale pretraining and contrastive objectives, where vision transformers \citep{dosovitskiy_image_2021} and CLIP \citep{radford_learning_2021} show strong near-OOD performance and benefit markedly from few-shot outlier exposure and even label-only supervision for outlier classes \citep{10.5555/3540261.3540802}. However, full fine-tuning can distort pretrained representations and harm OOD generalization relative to linear probing, with similar cautions for vision–language models; recent work also explores training-time scaling and post-hoc enhancements, and revisits detector design in vision foundation models \citep{10.5555/3540261.3540802,ming_how_2024,xu_scaling_2023,zhao_rethinking_2024}. Evaluation rigor has improved through ImageNet-scale suites like NINCO that mitigate in-distribution leakage. Meanwhile, theory and diagnostics connect feature separability to OOD error and delineate when OOD detection is learnable \citep{bitterwolf_or_2023,xieImportanceFeatureSeparability}.

Representation geometry and normalization have attracted increased attention for their role in OOD generalization. An analysis of contrastive learning and normalization approaches~\citep{le-gia_understanding_2023, tan_exploiting_2025} shows that geometric priors, such as hyperspherical projection or $\ell_2$ normalization, can yield more robust representation spaces. Studies like~\citep{zhao_towards_2024} and~\citep{xieImportanceFeatureSeparability} link improved feature separability and lower intrinsic dimensionality to higher OOD detection performance.

\begin{figure*}[t!]
    \centering
    \includegraphics[width=0.99\textwidth]{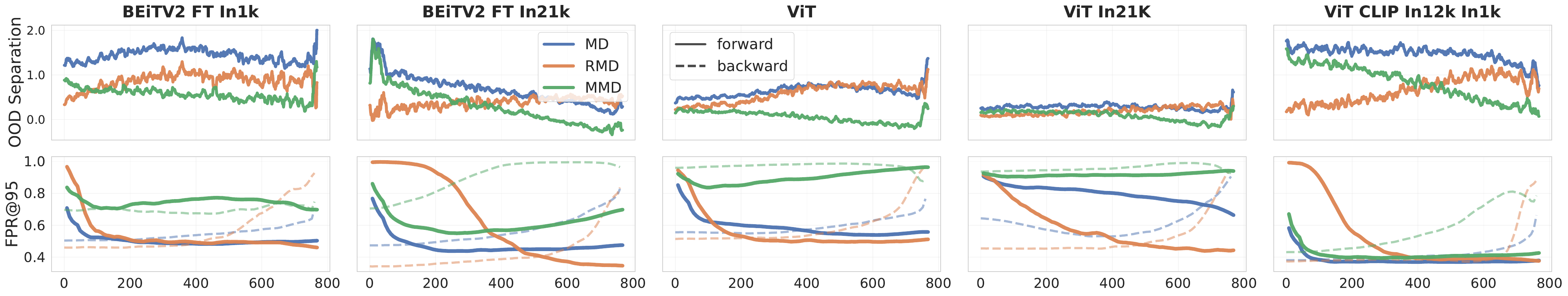}
    \caption{
    \textbf{Dimension-wise OOD behavior for Mahalanobis variants.} The top row shows dimension-wise OOD separation $S_i$, and the bottom row reports FPR under progressive dimension ablation. Strong embedding-space separation does not necessarily translate to better OOD detection.
    }
    \label{fig:combined_mds_analysis}
    \vspace{-0.5em}
\end{figure*}

\section{Background: Mahalanobis OOD Detection}
\label{sec:mahalanobis_theory}

Let $z=f(x)\in\mathbb{R}^d$ denote the feature representation of an input $x$.
Given ID training features from $K$ classes, Mahalanobis-based detectors model each class by a Gaussian
$\mathcal{N}(\mu_k,\Sigma)$ with a \emph{tied} covariance $\Sigma$ (the LDA assumption), and optionally a
marginal Gaussian $\mathcal{N}(\mu_0,\Sigma_0)$ fitted to all ID features.

\paragraph{Mahalanobis variants.}
The class-conditional Mahalanobis distance (MD) and its confidence score are
\begin{equation}
\begin{aligned}
\operatorname{MD}_k(z) &= (z-\mu_k)^\top \Sigma^{-1}(z-\mu_k),\\
\mathcal{C}_{\mathrm{MD}}(x) &= -\min_{k}\operatorname{MD}_k\!\big(f(x)\big).
\end{aligned}
\label{eq:md_def}
\end{equation}
Marginal Mahalanobis (MMD) uses the class-agnostic quadratic form
$\operatorname{MD}_0(z)=(z-\mu_0)^\top\Sigma_0^{-1}(z-\mu_0)$ as its score, and Relative Mahalanobis (RMD) \citep{ren_simple_2021}
subtracts this marginal reference:
\begin{equation}
\begin{aligned}
\operatorname{RMD}_k(z) &= \operatorname{MD}_k(z)-\operatorname{MD}_0(z),\\
\mathcal{C}_{\mathrm{RMD}}(x) &= -\min_k \operatorname{RMD}_k(z).
\end{aligned}
\label{eq:rmd_def}
\end{equation}

\paragraph{Eigenbasis view.}
Let $\Sigma=U\Lambda U^\top$ have eigenvalues $\lambda_1\ge\cdots\ge\lambda_d>0$ and eigenvectors $\{u_i\}$.
Define the per-direction energy $\tilde a_i(z)\triangleq (u_i^\top(z-\mu_k))^2$.
Then
\begin{equation}
\operatorname{MD}_k(z)=\sum_{i=1}^d \lambda_i^{-1}\,\tilde a_i(z).
\label{eq:md-eigen}
\end{equation}

This form highlights how directions with smaller variance (smaller $\lambda_i$) receive larger inverse weighting and motivates our later analysis of how representation geometry influences Mahalanobis-style OOD behavior.

\section{Comparative Study of Self-Supervised Models}

\subsection{Cross-model OOD Detection Performance}
\label{sec:cross_model_ood_perf}
We begin by characterizing how representation learning choices shape Mahalanobis-style OOD detection across modern vision backbones. In particular, we ask how performance depends on architecture, pretraining data, and fine-tuning regime, i.e., factors whose effects are not systematically documented. This motivates a broad, model-agnostic comparison:
\emph{Which modern self-supervised or pretrained vision models produce representations that naturally lend themselves to Mahalanobis-style OOD detection?}

\paragraph{Models.}
We evaluate publicly available checkpoints from \texttt{timm}~\citep{rw2019timm} and \texttt{huggingface-transformers}~\citep{wolf-etal-2020-transformers}, covering multiple transformer families, model scales, and training objectives. The full model list is provided in Appendix \ref{app:full_model_names}.

\paragraph{Evaluation protocol.}
Following OpenOOD~\citep{10.5555/3600270.3602632}, ImageNet-1K serves as the in-distribution (ID) dataset (train features for fitting; validation for ID testing). We report $\mathrm{FPR}@95$ for distinguishing ImageNet validation from each of the five OOD benchmarks: NINCO~\citep{bitterwolf_or_2023}, iNaturalist~\citep{Horn_2018_CVPR}, SSB-Hard~\citep{bitterwolf_or_2023}, OpenImages-O~\citep{OpenImages2}, and Textures~\citep{Cimpoi_2014_CVPR}. Unless stated otherwise, we fit class means $\{\mu_k\}$ and a tied covariance $\Sigma$ on ImageNet-1K training features and evaluate OOD scores on the ImageNet validation set versus each OOD dataset.

\begin{figure}[h]
    \centering
    \includegraphics[width=\columnwidth]{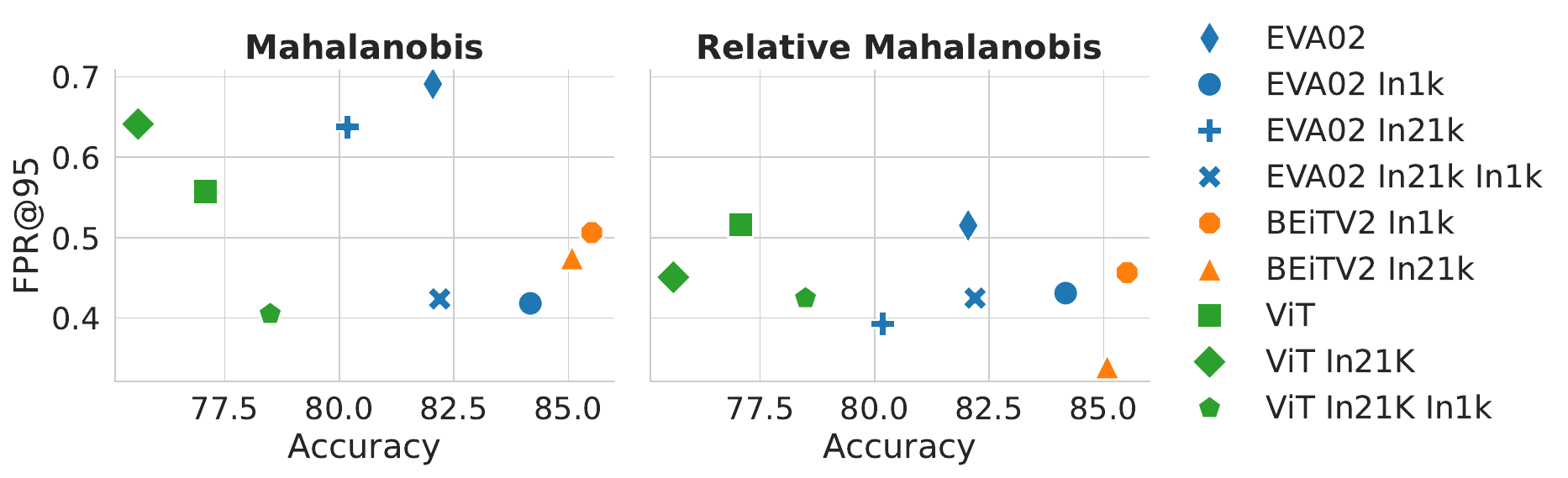}
    \caption{
    \textbf{OOD detection performance on NINCO across model families.} RMDS consistently improves over standard MD, with the largest gains for models that are pretrained but not fine-tuned on ImageNet.
    }
    \label{fig:ood_study_mds_ninco_compact}
\end{figure}

\paragraph{Results (Figure~\ref{fig:ood_study_mds_ninco_compact}).}
RMD improves over the standard Mahalanobis distance in most settings, with the largest gains for models that are pretrained but not fine-tuned on ImageNet.
Notably, RMD substantially improves OOD detection for EVA02-In21k and ViT-In21k, in some cases matching or exceeding the performance of their ImageNet-fine-tuned counterparts.
This weakens the typical association between in-distribution accuracy and FPR, and produces more consistent score distributions across models.
At the same time, classification accuracy is not a reliable proxy for OOD performance: large accuracy gaps (often $>10\%$) do not necessarily yield better detection.
We observe only a mild trend along the fine-tuning sequence In1k $\rightarrow$ In22k-In1k $\rightarrow$ larger In22k-In1k models; full results are reported in Appendix~\ref{app:cross_model_performance}.

\begin{figure*}[t]
    \centering
    \includegraphics[width=0.7\textwidth]{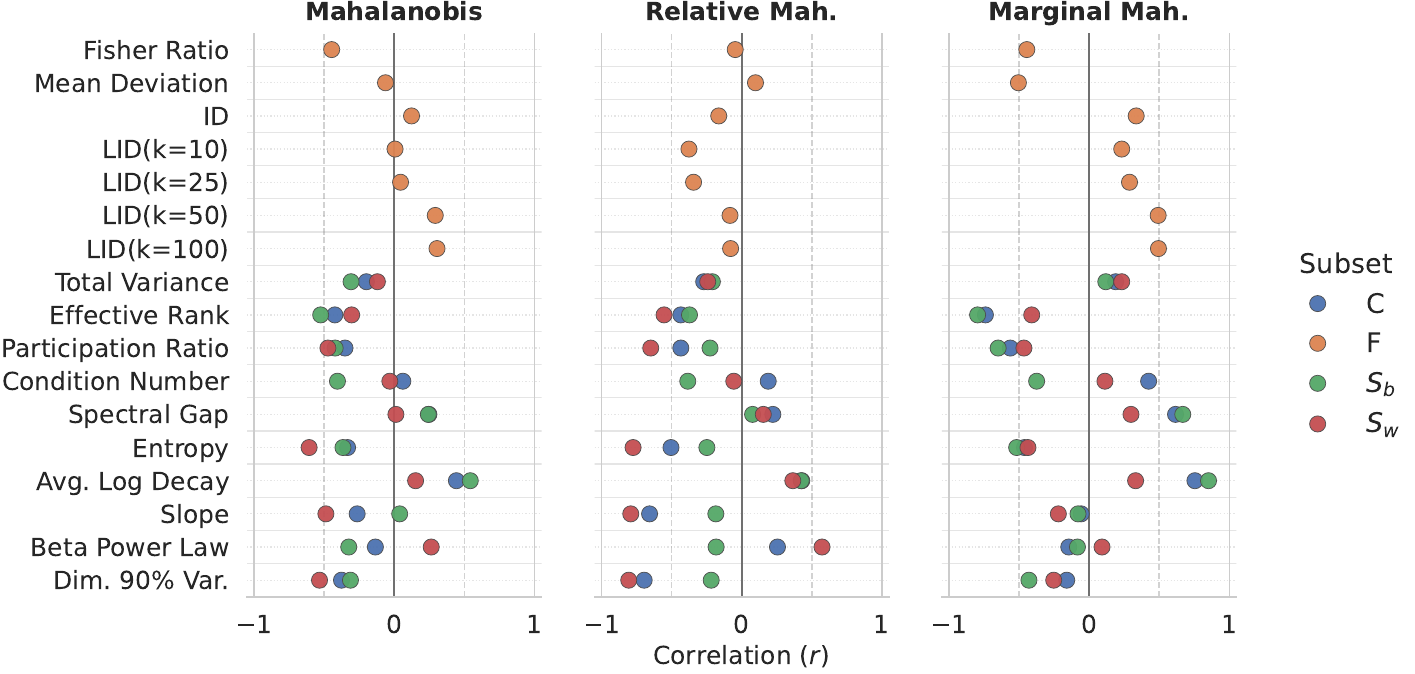}
    \caption{
    \textbf{Correlations between representation metrics and OOD performance.}
    Spearman correlations between representation metrics and OOD performance across Mahalanobis variants. Here, \textbf{F} denotes metrics computed directly from feature representations, whereas \textbf{C}, $\mathbf{S_b}$, and $\mathbf{S_w}$ are derived from the eigenvalue spectra of the covariance/scatter matrices (Appendix~\ref{app:metrics}). The three Mahalanobis-based detectors leverage distinct geometric cues, yielding different correlation patterns. Similar trends hold for Pearson correlations (Appendix~\ref{app:pearson-correlation}).
    }
    \label{fig:correlation_heatmap_spearman}
    \vspace{-0.5em}
\end{figure*}

\subsection{Mahalanobis Variants and Per-Dimension Analysis}
\label{sec:md-per-dim}
Having established cross-model trends, we next ask:
\emph{which parts of the representation space drive OOD discrimination?}
Beyond aggregate scores, inspecting individual eigen-directions helps explain why some models detect OOD reliably while others do not. We therefore analyze MD, MMD, and RMD at the level of their \emph{per-direction} contributions. 

\paragraph{Per-direction separation.}
Using the decomposition in Eq.~\ref{eq:md-eigen}, we define the OOD separation in the eigen-direction $i$ as the difference between its mean contribution to OOD and ID:
\begin{equation}
S_i \;\triangleq\; \lambda_i^{-1}\!\left(
\mathbb{E}_{x\sim\mathcal{D}_{\mathrm{OOD}}}[\tilde a_i(x)]
-\mathbb{E}_{x\sim\mathcal{D}_{\mathrm{ID}}}[\tilde a_i(x)]
\right),
\label{eq:Si_def}
\end{equation}
where the projection term $\tilde a_i(\cdot)$ utilizes the selected class $k \in \arg\min_c \operatorname{MD}_c$ (or $\arg\min_c \operatorname{RMD}_c$ for RMD; see Eq. \ref{eq:md_def} and \ref{eq:rmd_def}).
$S_i$ measures the difference between average ID and OOD Mahalanobis energy along eigen-direction $i$. 
Positive $S_i$ indicates that OOD samples contribute more inverse-variance weighted energy than ID samples along $u_i$. We order eigenvalues as $\lambda_1\ge\cdots\ge\lambda_d$, so larger $i$ corresponds to smaller-variance directions.

\paragraph{Ablation protocol.}
Figure~\ref{fig:combined_mds_analysis} (bottom) reports an ablation study where we recompute the FPR using only the first $K$ eigen-directions (forward ablation) or only the last $K$ directions (backward ablation).
This isolates whether discrimination arises from high-variance structure (small $i$) or from low-variance components that receive strong inverse-variance weighting (large $i$).

\paragraph{Results.}
Figure~\ref{fig:combined_mds_analysis} reveals three recurring behaviors across models.
(i) Large per-direction separation does not necessarily yield low FPR: for example, BEiTV2 FT In1k exhibits stronger separation across many directions yet performs on par with (or worse than) BEiTV2 FT In21k.
(ii) The effective number of eigen-directions needed for strong detection varies substantially: some models saturate quickly with small $K$, while others require most of the spectrum to approach their best FPR.
(iii) Backward ablation shows that low-variance directions can dominate discrimination in some settings; e.g., ViT In21k attains its best FPR primarily from the latter part of the spectrum, suggesting that small-variance components can carry disproportionate OOD signal after inverse-variance weighting.
These observations motivate our later stability analysis: performance is governed not only by mean ID–OOD separation but also by how quadratic weighting interacts with the representation spectrum and the allocation of sample energy across eigen-directions.

\section{Geometry of Representations}
In the previous section, we showed that no single OOD method yields consistent performance and behavior across multiple models. In fact, different SSL models and pretraining regimes produce representations with distinct geometric properties, indicating that OOD performance depends on the intrinsic structure of the representation space. To understand these effects, we analyze the internal geometry of model representations, seeking to answer: \emph{What internal characteristics of a model’s feature space predict strong OOD detection?}

\subsection{Geometry-detector alignment}
\label{sec:geometric_trade_offs}
To identify which representation properties matter for OOD detection, we correlate detection performance with two complementary families of metrics:
(i) \textbf{manifold metrics}, such as intrinsic dimensionality ~\citep{ma_characterizing_2018}, computed from the ID features $F$, 
and (ii) \textbf{spectral metrics} computed from the eigenspectra of the global covariance $C$ and Fisher scatter matrices $S_w,S_b$
(see Appendix~\ref{app:metrics} for definitions).
Figure~\ref{fig:correlation_heatmap_spearman} summarizes Spearman correlations across detectors and models.

As we can see, the RMD correlates most strongly with within-class geometry ($S_w$), reflecting the importance of compact, well-structured class clusters. MMD correlates primarily with global geometry ($C$ and $S_b$), indicating dependence on the overall manifold shape. Standard MD sits between these extremes, combining sensitivity to both cluster structure and the global eigenspectrum. In Appendix \ref{app:spectral_analysis}, we provide a spectral analysis that offers insight into how pretraining and fine-tuning shape these characteristics.

\subsection{Ideal geometry: a compensatory trade-off}
\label{sec:ideal_geometry}

\begin{figure}[h]
    \centering
    \includegraphics[width=0.99\linewidth]{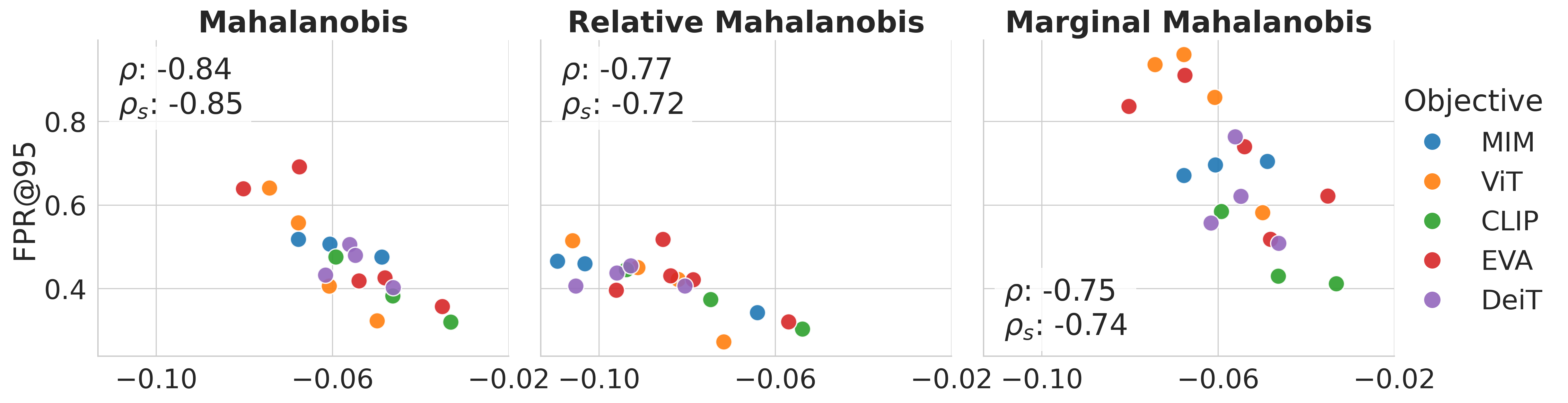}
    \caption{\textbf{A simple predictor of Mahalanobis OOD performance.} The product $m|s|$ (LID $\times$ within-class spectral-slope magnitude) correlates with Mahalanobis-based OOD detection across variants; lower values indicate better performance.}
    \label{fig:metric_combination_correlation}
\end{figure}

OOD detection depends not only on mean ID-OOD separation but also on how \emph{ID variability is organized}. Two geometric factors repeatedly appear across models:
\textbf{local degrees of freedom} (how many directions are explored in a neighborhood) and \textbf{within-class concentration} (how tightly class clusters concentrate around their means). Intuitively, if the local manifold is simple (low LID), reliable detection requires very compact clusters; if the local manifold is richer (high LID), OOD samples can deviate along many directions, allowing for less compact clusters. This yields a compensatory trade-off between local dimension and within-class concentration. Empirically, this trade-off is captured by the product of LID ($m$) and the within-class spectral slope ($|s|$) magnitude. Figure~\ref{fig:metric_combination_correlation} shows that $m \cdot |s|$ (LID $\times$ slope of $S_w$ spectrum) strongly predicts Mahalanobis OOD performance across models and variants. We estimate the dataset-average LID $m$ using the $k$NN LID estimator with $k=50$ and keep $k$ fixed across all experiments for consistency. The product $m|s|$ is expected to be stable to small variations in $k$ within the local regime (see Table~\ref{tab:ust_proxy_tracking} in Appendix~\ref{app:ust_proxy}).

Taken together, these results suggest that neither local dimensionality nor within-class concentration alone is sufficient to explain Mahalanobis OOD behavior. Instead, strong performance is associated with representations that jointly balance these two properties, for which $m|s|$ provides a compact summary. 
This suggests that if we can move a representation along the $m \times |s|$ trade-off curve post-hoc, we may tune Mahalanobis behavior without changing the backbone. In the following section, we show how this summary can be \emph{traced and optimized} by a simple post-hoc deformation of the feature space and later provide a mechanistic explanation for why $m|s|$ is predictive across Mahalanobis variants. 

\section{Radial scaling as a geometric control knob}
\label{sec:normalization_effects}

The preceding section identifies $m|s|$ as a detector-invariant summary of ID geometry that predicts Mahalanobis OOD performance.
We now introduce a simple post-hoc mechanism to \emph{modify} this geometry without changing the backbone: a one-parameter family of
\textbf{direction-preserving radial deformations}.
This family generalizes standard $\ell_2$ normalization (unit-sphere projection) and provides a continuous knob for navigating a trajectory
$\beta \mapsto (m(\beta), s(\beta))$ using \emph{ID data only}.

\subsection{Why radial normalization?}
\label{sec:why_radial}

Mahalanobis-style detectors are quadratic forms that depend on a shared covariance estimate.
In practice, feature norms can vary substantially across samples and models, and this radial variability can dominate the covariance fit and inflate score overlap between ID and OOD.
Prior work has shown that $\ell_2$ normalizing features (projecting to the unit sphere) can stabilize Mahalanobis OOD detection by reducing norm-driven variation and improving the fit of quadratic scores \citep{mueller_mahalanobis_2025}.
We choose a direction-preserving radial family because it is the minimal intervention that changes norm-driven geometry while preserving angular class structure.

\begin{table*}[t]
\centering
\caption{\textbf{OOD detection (FPR@95, $\downarrow$)} averaged over five OpenOOD datasets.
\textit{MD} uses features as-is ($\beta=0$), \textit{MD++} applies $\ell_2$ normalization ($\beta=1$).
\textit{RS-MD} selects a per-model radial exponent $\hat{\beta}$ from ID data by selecting optimal geometric proxy
$P(\beta)=m(\beta)\lvert s(\beta)\rvert$ over the search grid.
\textit{RMD} denotes the relative Mahalanobis variant, with analogous \textit{RMD++} ($\beta=1$) and \textit{RS-RMD} (proxy-selected $\hat{\beta}$) settings. For each model (column), \textit{RS-MD} is highlighted in \textcolor{lightgreen}{light green} when it improves over both \textit{MD} and \textit{MD++}; likewise, \textit{RS-RMD} when it outperforms both \textit{RMD} and \textit{RMD++}.
Best (\textbf{lowest}) results in each column are bolded.}
\label{tab:main_results_fpr}
\adjustbox{max width=\textwidth}{
  \begin{tabular}{lcccccccccccc|c}
\toprule
Detector &
\makecell{BEiTv2\\In1k} &
\makecell{ViT\\In1k} &
\makecell{ViT\\In21k$\rightarrow$In1k} &
\makecell{ViT-L\\In21k$\rightarrow$In1k} &
\makecell{DeiT3\\In1k} &
\makecell{DeiT3\\In21k$\rightarrow$In1k} &
\makecell{DeiT3-L\\In22k$\rightarrow$In1k} &
\makecell{EVA02\\In1k} &
\makecell{EVA02\\In21k$\rightarrow$In1k} &
\makecell{CLIP\\In1k} &
\makecell{CLIP\\In12k$\rightarrow$In1k} &
\makecell{CLIP-L\\In12k$\rightarrow$In1k} &
Avg \\
\midrule
MSP & 52.2 & 56.5 & 53.7 & 44.8 & 55.0 & 56.7 & 58.1 & 53.2 & 53.0 & 55.2 & 49.0 & 45.0 & 52.7 \\
MLS & 50.7 & 50.4 & 40.7 & 29.8 & 59.2 & 64.3 & 65.9 & 55.3 & 58.9 & 65.3 & 51.9 & 43.6 & 53.0 \\
KNN & 42.6 & 50.0 & 47.7 & 34.3 & 47.5 & 37.0 & 35.9 & 40.6 & 42.3 & 41.1 & 32.5 & 30.1 & 40.1 \\
VIM & 39.3 & 53.0 & 36.1 & \bfseries 25.0 & 47.2 & 37.5 & 39.7 & 43.9 & \bfseries 37.1 & 41.7 & 30.0 & 28.0 & 38.2 \\
RS-MD & {\cellcolor{lightgreen}} \bfseries 37.2 & 45.5 & 35.8 & 25.3 & 43.2 & {\cellcolor{lightgreen}} \bfseries 35.5 & {\cellcolor{lightgreen}} \bfseries \bfseries 33.4 & {\cellcolor{lightgreen}} \bfseries \bfseries 37.2 & 39.5 & {\cellcolor{lightgreen}} \bfseries \bfseries 37.6 & {\cellcolor{lightgreen}} \bfseries \bfseries 26.4 & {\cellcolor{lightgreen}} \bfseries \bfseries 26.7 & {\cellcolor{lightgreen}} \bfseries \bfseries 35.3 \\
MD++ & \color{gray} 37.6 & 45.4 & 38.7 & 28.2 & 43.0 & \color{gray} 35.6 & \color{gray} 34.2 & \color{gray} 37.4 & 38.2 & \color{gray} 38.2 & \color{gray} 27.8 & \color{gray} 27.1 & \color{gray} 36.0 \\
MD & \color{gray} 40.2 & 45.7 & \bfseries 35.7 & 25.3 & 43.3 & \color{gray} 37.6 & \color{gray} 36.6 & \color{gray} 37.6 & 40.8 & \color{gray} 40.2 & \color{gray} 33.5 & \color{gray} 29.7 & \color{gray} 37.2 \\
RS-RMD & {\cellcolor{lightgreen}} \bfseries \bfseries 37.1 & 44.8 & {\cellcolor{lightgreen}} \bfseries 37.5 & 26.9 & {\cellcolor{lightgreen}} \bfseries \bfseries 39.4 & {\cellcolor{lightgreen}} \bfseries \bfseries 34.5 & {\cellcolor{lightgreen}} \bfseries 35.5 & {\cellcolor{lightgreen}} \bfseries 39.6 & {\cellcolor{lightgreen}} \bfseries 38.6 & {\cellcolor{lightgreen}} \bfseries 38.2 & {\cellcolor{lightgreen}} \bfseries 30.7 & {\cellcolor{lightgreen}} \bfseries 27.0 & {\cellcolor{lightgreen}} \bfseries 35.8 \\
RMD++ & \color{gray} 37.3 & \bfseries 44.6 & \color{gray} 37.6 & 26.9 & \color{gray} 39.9 & \color{gray} 35.1 & \color{gray} 35.9 & \color{gray} 39.8 & \color{gray} 39.1 & \color{gray} 38.6 & \color{gray} 30.9 & \color{gray} 27.7 & \color{gray} 36.1 \\
RMD & \color{gray} 39.1 & 44.9 & \color{gray} 37.6 & 26.9 & \color{gray} 40.8 & \color{gray} 36.6 & \color{gray} 37.6 & \color{gray} 40.3 & \color{gray} 40.3 & \color{gray} 40.3 & \color{gray} 32.5 & \color{gray} 29.3 & \color{gray} 37.2 \\
\bottomrule
\end{tabular}

}
\vspace{-0.5em}
\end{table*}

\subsection{Radially scaled Mahalanobis distance}
\label{sec:radial_l2}

Given a feature vector $z \in \mathbb{R}^d \setminus \{0\}$, we define the radial map
\begin{equation}
\phi_\beta(z) \;=\; \frac{z}{\|z\|^{\beta}},
\label{eq:radial_l2_norm}
\end{equation}
where $\beta \in \mathbb{R}$ controls radial contraction/expansion while preserving direction.
The induced radius is $\|\phi_\beta(z)\|=\|z\|^{1-\beta}$. Thus, for $\beta>1$, the map contracts norms greater than 1 and expands norms smaller than 1 (pushing toward the unit sphere); $\beta<1$ has the opposite tendency (pushing away from the unit sphere); and $\beta<0$ expands the norms (see Appendix~\ref{app:radial-metric} for more details). 
This family contains key special cases:
$\beta=0$ recovers the original geometry, and $\beta=1$ projects features onto the unit sphere. In practice, most feature norms are $>1$. 

\paragraph{Definition.}
We denote by \textbf{RS-MD} the Mahalanobis-distance detector applied to features after the transformation $\phi_\beta$ in Eq.~\ref{eq:radial_l2_norm}. 
Thus, $\beta=0$ recovers standard MD, and $\beta=1$ corresponds to the $\ell_2$-normalized variant (MD++).
We define \textbf{RS-RMD} analogously for the relative Mahalanobis variant evaluated on $\phi_\beta(z)$.

\subsection{Predicting the Optimal \texorpdfstring{$\beta$}{beta}}

\paragraph{What Changes Under $\phi_\beta$.}
Applying $\phi_\beta$ modifies both (i) local neighborhood structure (and thus LID $m(\beta)$) and
(ii) the spectrum of within-class scatter $S_w(\beta)$ (and thus the slope $s(\beta)$). Thus, changing $\beta$ systematically alters (i) neighborhood geometry (captured by $m(\beta)$) and (ii) within-class spectral decay (captured by $s(\beta)$), which in turn shifts the quadratic scores through the refit scatter estimate.

\begin{figure}[t]
    \centering
    \includegraphics[width=0.8\columnwidth]{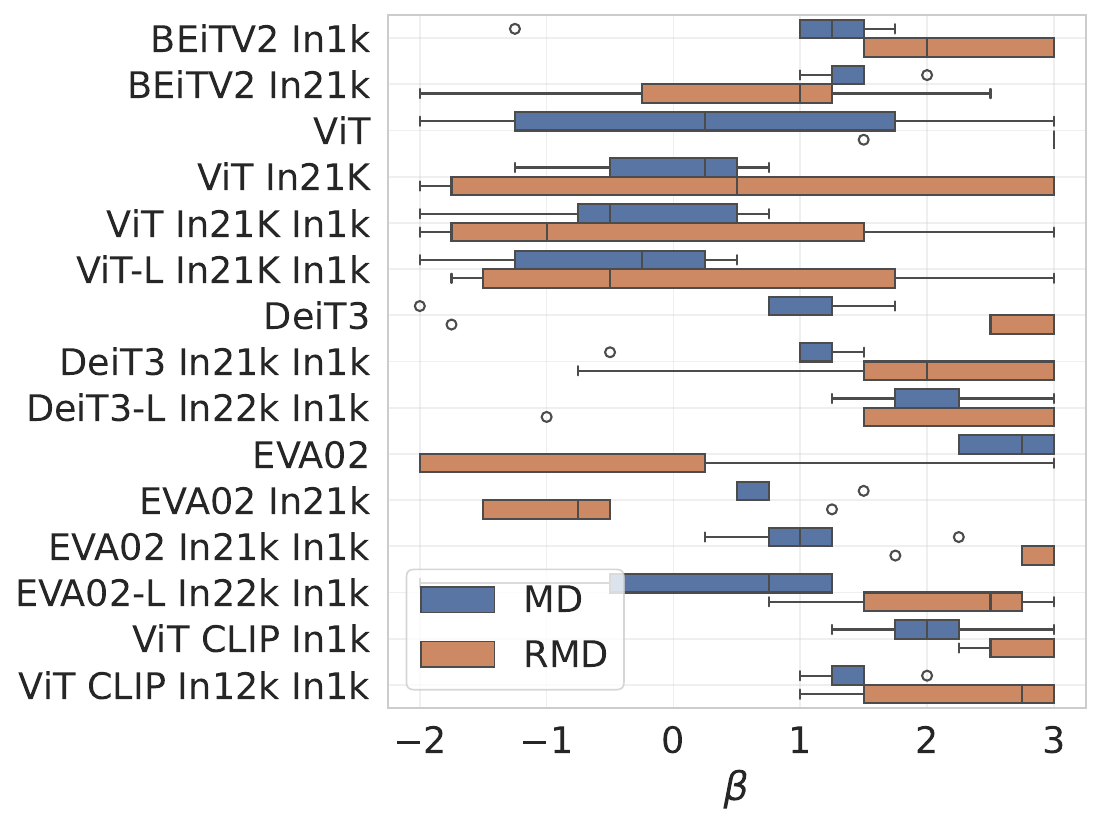}
    \caption{
    \textbf{Empirically optimal $\beta$ varies across OOD settings.}
    Distribution of the empirically optimal $\beta$ for MD and RMD detectors across OOD datasets. The wide spread highlights substantial model- and dataset-specific variation, indicating that $\beta$ typically requires tuning per setting.
    }
    \label{fig:optimal_beta_per_model_boxplot}
\end{figure}

Our experiments show that the optimal radial parameter $\beta$ is highly model- and dataset-dependent, reflecting the intrinsic geometry of the learned representations. Moderate positive values often align the feature distribution with the Gaussian, tied-covariance assumptions of the Mahalanobis detector; yet in some cases, larger or even negative values yield stronger in/out-of-distribution separation. Consequently, a fixed choice of $\beta$ is rarely optimal. Figure~\ref{fig:optimal_beta_per_model_boxplot} shows the empirically optimal $\beta$ values (searched over $[-2,2]$ in $0.25$ steps) for MD and RMD detectors across different OOD datasets. The wide spread of optimal values underscores that a one-size-fits-all approach is ineffective.

\subsection{ID-only selection of $\beta$ via Geometric Proxy}
\label{sec:beta_proxy_selection}

\paragraph{Setup.}
We study ID-only selection of $\beta$ on ImageNet-trained transformer backbones spanning several families (BEiT-v2, CLIP, EVA, and ViT variants with different pre-training and fine-tuning regimes). For each backbone and each $\beta$ on a discrete grid $\mathcal{B}\subset[-2,2]$, we estimate two quantities on a held-out portion of the ImageNet-1K training data: (i) the dataset-average local intrinsic dimensionality (LID) $m(\beta)$ using a $k{=}50$ nearest-neighbor estimator (as in Section~\ref{sec:ideal_geometry}), and (ii) the within-class spectral slope $s(\beta)$ obtained from the eigenspectrum of the within-class scatter matrix $S_w(\beta)$ computed in the feature space induced by $\phi_\beta(z)$.
OOD detection is then evaluated using the same datasets, protocol, and metrics as in Section~\ref{sec:cross_model_ood_perf}; unless stated otherwise, the selected $\hat{\beta}$ is applied unchanged for that model on ImageNet (ID) and on all OOD benchmarks.

\paragraph{Proxy definition}
For each $\beta$, we compute (i) the dataset-average LID $m(\beta)$ and (ii) the within-class spectral slope $s(\beta)$ from the eigenspectrum of $S_w(\beta)$ formed on $\phi_\beta(z)$. We combine them into a single ID-only geometry summary,
\begin{equation}
P(\beta) \triangleq m(\beta)\,|s(\beta)|,
\label{eq:proxy_curve}
\end{equation}
and evaluate $P(\beta)$ over the grid $\mathcal{B}\subset[-2,2]$,

\paragraph{Selecting $\hat{\beta}$ from the proxy curve}
Across models, $P(\beta)=m(\beta)|s(\beta)|$ typically has an interior turning point (often inverted-U; occasionally U-shaped, e.g., ViTs). Since boundary optima can be artifacts of the finite search range, we select the most pronounced \emph{interior} turning point: the interior grid value farthest from the endpoint baseline $P_{\mathrm{end}}=(P(\beta_{\min})+P(\beta_{\max}))/2$ (Appendix~\ref{app:beta_selection_rule}). This recovers the interior maximum for inverted-U curves and the interior minimum for U-shaped curves; if no clear interior turning point appears on the grid, we use the Appendix fallback. In Section~\ref{sec:mahalanobis_theory} connect these regimes to the instability functionals $\mathcal I(\beta)$ and $\widehat{\mathcal I}(\beta)$. Note that Section~\ref{sec:ideal_geometry} reports an across-model correlation at $\beta=0$, which need not determine the within-model optimum along $\beta$.

\begin{figure}[t]
    \centering
    \includegraphics[width=0.95\columnwidth]{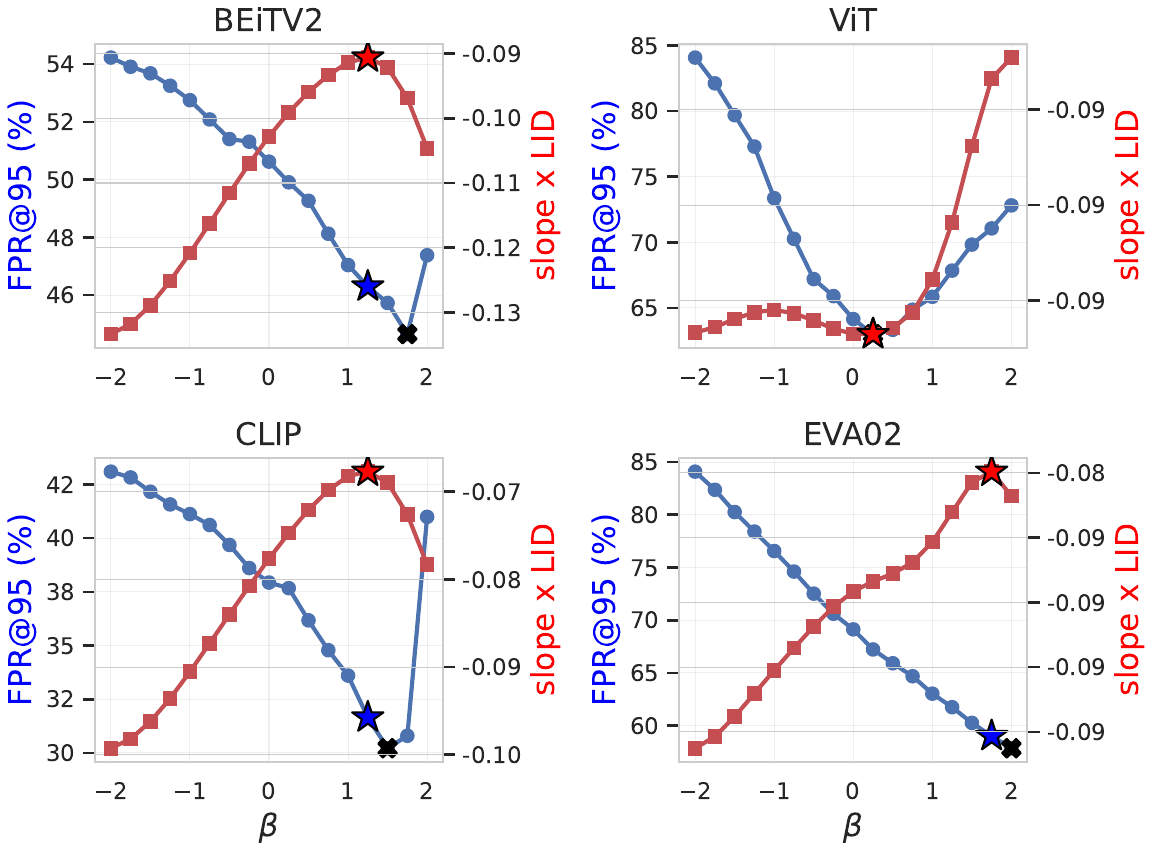}
    \caption{\textbf{Proxy feature–FPR trade-off across $\beta$ on NINCO (standard MD).} Each panel corresponds to a model configuration. The \textcolor{blue}{blue} curve reports the \textcolor{blue}{$\mathrm{FPR}@95$} as a function of $\beta$, while the \textcolor{red}{red} curve (right axis) shows the in-distribution proxy feature \textcolor{red}{$P(\beta)\triangleq m(\beta)|s(\beta)|$}. The star indicates the $\beta$ selected by the proxy, and the black \textbf{x} denotes the oracle $\beta$ that minimizes $\mathrm{FPR}@95$.}
    \label{fig:slope_x_lid_tuning}
    \vspace{-0.8em}
\end{figure}

\begin{figure}[h]
    \centering
    \vspace{-0.5em}
    \includegraphics[width=0.7\columnwidth]{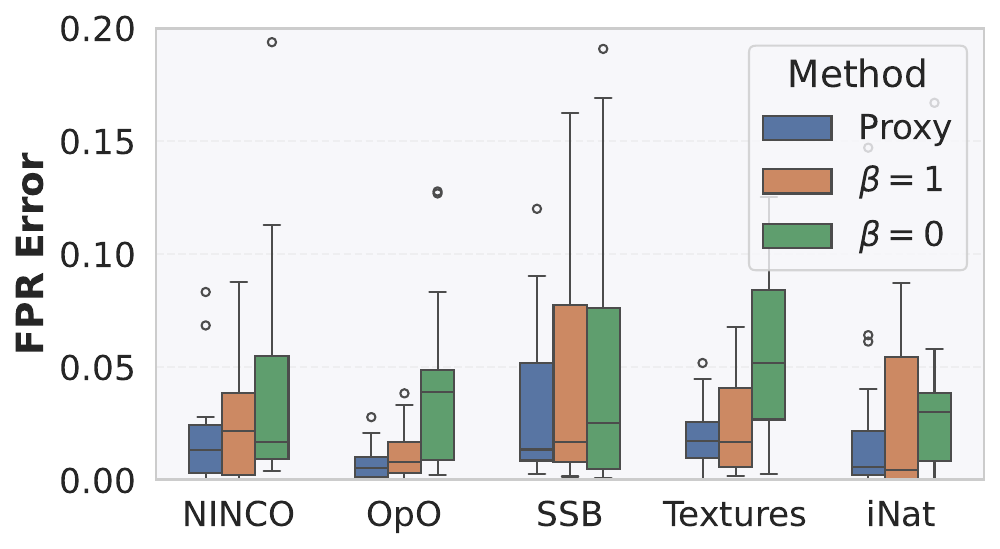}
    \caption{\textbf{Absolute FPR error relative to the oracle $\beta^*$ (standard MD).} Boxplots show $|\mathrm{FPR}(\hat{\beta})-\mathrm{FPR}(\beta^*)|$ across OOD benchmarks. The proxy-selected $\hat{\beta}$ consistently achieves lower error than the fixed baselines $\beta{=}0$ and $\beta{=}1$ on every dataset.}
    \label{fig:predictions_beta_md_proxy}
    \vspace{-0.5em}
\end{figure}

\paragraph{OOD performance summary}
Table~\ref{tab:main_results_fpr} reports $\mathrm{FPR}@95$ across backbones and OOD datasets. The proxy-selected $\hat{\beta}$ consistently improves OOD detection relative to fixed choices (e.g., $\beta=0$ for standard MD and $\beta=1$ for MD++), for both MD and RMD. Although the proxy does not always match the empirically optimal $\beta$, it captures enough ID geometric structure to yield meaningful gains in detection performance. 
Figure~\ref{fig:predictions_beta_md_proxy} further quantifies selection quality for standard MD by plotting the absolute error to the oracle choice $\beta$, i.e., $|\mathrm{FPR}(\hat{\beta})-\mathrm{FPR}(\beta)|$, across OOD benchmarks and the same set of models as in Table~\ref{tab:main_results_fpr}. The proxy achieves lower error than the $\beta=0$ and $\beta=1$ baselines across datasets, with strong improvements on NINCO, whose samples were verified to be free of ID contamination. The proxy also reduces worst-case behavior, as reflected by a lower upper tail (fewer/lower outliers) in the error distribution.
Full results and comparisons against baseline OOD detectors are provided in Appendix~\ref{app:detailed-ood-performance}.

\section{A Unified Stability Lens for Mahalanobis Variants}
\label{sec:ust}

The previous sections establish two empirical facts: (i) Mahalanobis-style OOD detection exhibits large cross-model variability, and
(ii) the composite geometry summary $m(\beta)\,|s(\beta)|$ tracks performance across models and along $\beta$-trajectories.
We now provide a mechanistic lens that explains \emph{why} a low-dimensional ID-only geometry summary can be predictive while also clarifying its limitations.

\subsection{Exact size--stretch factorization of quadratic scores}
\label{sec:ust_factorization}
Let $\delta_\beta(z)\in\mathbb{R}^d$ denote the detector-specific centered deviation in $\phi_\beta$-space (e.g., class-conditional for MD, global for MMD),
and let $\Sigma(\beta)\in\mathbb{R}^{d\times d}$ be the corresponding tied ID scatter/covariance used by the detector.
We assume $\Sigma(\beta)\succ 0$ for all $\beta$ under consideration, so that $\Sigma(\beta)^{-1}$ exists.\footnote{In practice, we enforce invertibility via standard regularization.}
For any Mahalanobis-type quadratic score
\begin{equation}
S_\beta(z)\;=\;\delta_\beta(z)^\top \Sigma(\beta)^{-1}\delta_\beta(z),
\label{eq:S_def}
\end{equation}
define the \emph{whitened stretch factor}
\begin{equation}
W_\beta(z)\;\triangleq\;\frac{\delta_\beta(z)^\top \Sigma(\beta)^{-1}\delta_\beta(z)}{\|\delta_\beta(z)\|^2},
\label{eq:W_def}
\end{equation}
which is well-defined whenever $\|\delta_\beta(z)\|>0$.\footnote{We assume $\|\delta_\beta(z)\|>0$ almost surely under $\mathcal D_{\mathrm{ID}}$; empirically, exact zeros occur with negligible frequency and are dropped.}
Then the score admits the exact factorization
\begin{equation}
S_\beta(z)=\|\delta_\beta(z)\|^2\,W_\beta(z)
\label{eq:factorization}
\end{equation}
and hence $\log S_\beta(z)=\log\|\delta_\beta(z)\|^2+\log W_\beta(z)$.
This separates score variability into a \emph{size} channel ($\|\delta\|^2$) and a \emph{stretch} channel ($W$) that depends on alignment with the whitening geometry $\Sigma(\beta)^{-1}$.

\paragraph{Relative Mahalanobis Distance.}
Eq.~\eqref{eq:S_def} covers detectors whose score is a \emph{single} quadratic form (MD, MMD).
RMD is a \emph{difference} of two quadratic forms that generally use distinct scatter estimates,
$S_{\mathrm{RMD},\beta}(z)=S^{(1)}_\beta(z)-S^{(2)}_\beta(z)$, and therefore does not admit a single shared $\Sigma(\beta)$ in general.
Our stability lens applies to the RMD \emph{term-wise}: each component $S^{(j)}_\beta(z)=\delta^{(j)}_\beta(z)^\top \Sigma^{(j)}(\beta)^{-1}\delta^{(j)}_\beta(z)$ admits the same factorization and channel decomposition, and RMD behavior additionally depends on interactions between the two terms (e.g., cancelation via cross-covariances). We provide the precise term-wise formulation in Appendix~\ref{app:rmd_ust}.

\subsection{ID instability decomposes into size, stretch, and compensation}
\label{sec:ust_decomposition}

To quantify how variable the ID scores are at a fixed operating point, we study the instability functional
\begin{equation}
\mathcal I(\beta)\;\triangleq\;\operatorname{Var}_{z\sim\mathcal D_{\mathrm{ID}}}\!\big[\log S_\beta(z)\big].
\label{eq:I_def}
\end{equation}
Using Eq.~\eqref{eq:factorization} and the variance-covariance decomposition, $\mathcal I(\beta)$ decomposes exactly as 
\begin{equation}
\mathcal I(\beta) = A_\delta(\beta)+A_W(\beta)+2A_\times(\beta),
\label{eq:I_decomp}
\end{equation}
where
\begin{equation}
\label{eq:channels}
\begin{aligned}
A_\delta(\beta) &\triangleq \operatorname{Var}_{z\sim\mathcal{D}_{\mathrm{ID}}}\!\big[\log\|\delta_\beta(z)\|^2\big],\\
A_W(\beta) &\triangleq \operatorname{Var}_{z\sim\mathcal{D}_{\mathrm{ID}}}\!\big[\log W_\beta(z)\big],\\
A_\times(\beta) &\triangleq \operatorname{Cov}_{z\sim\mathcal{D}_{\mathrm{ID}}}\!\big(\log\|\delta_\beta(z)\|^2,\ \log W_\beta(z)\big).
\end{aligned}
\end{equation}
Empirically, both $A_\delta$ and $A_W$ vary substantially across models and across $\beta$, while $A_\times$ is typically negative, indicating systematic \emph{compensation} between size and stretch (Appendix~\ref{app:ust_defs}). Consequently, no single scalar geometry statistic should be expected to mediate performance universally.

\subsection{Stretch variability is driven by spectrum \(\times\) allocation geometry}
\label{sec:ust_stretch}

Let \(\Sigma(\beta)=U(\beta)\Lambda(\beta)U(\beta)^\top\) and define the allocation of the deviation onto the eigenbasis by
\begin{equation}
p_i(z;\beta)\triangleq\frac{(u_i(\beta)^\top\delta_\beta(z))^2}{\|\delta_\beta(z)\|^2},\qquad \sum_{i=1}^d p_i(z;\beta)=1.
\label{eq:p_def}
\end{equation}
Then the stretch factor admits the exact identity
\begin{equation}
W_\beta(z)=\sum_{i=1}^d\frac{p_i(z;\beta)}{\lambda_i(\beta)}.
\label{eq:W_eig}
\end{equation}
Equation~\eqref{eq:W_eig} makes explicit that \(A_W\) is controlled by the interaction of spectral heterogeneity
(\(\lambda_i(\beta)\)) and pointwise allocation geometry (\(p(z;\beta)\)).
We validate this mechanism with targeted interventions: rotations (changing allocations), allocation smoothing (collapsing allocation variability),
and spectrum shaping (changing eigenvalues), all of which causally affect \(A_W\) as predicted (Appendix~\ref{app:ust_interventions}).

\subsection{A compact ID-only proxy law along \(\beta\)}
\label{sec:ust_proxy}

\begin{figure}[t]
  \centering
  \includegraphics[width=0.7\linewidth]{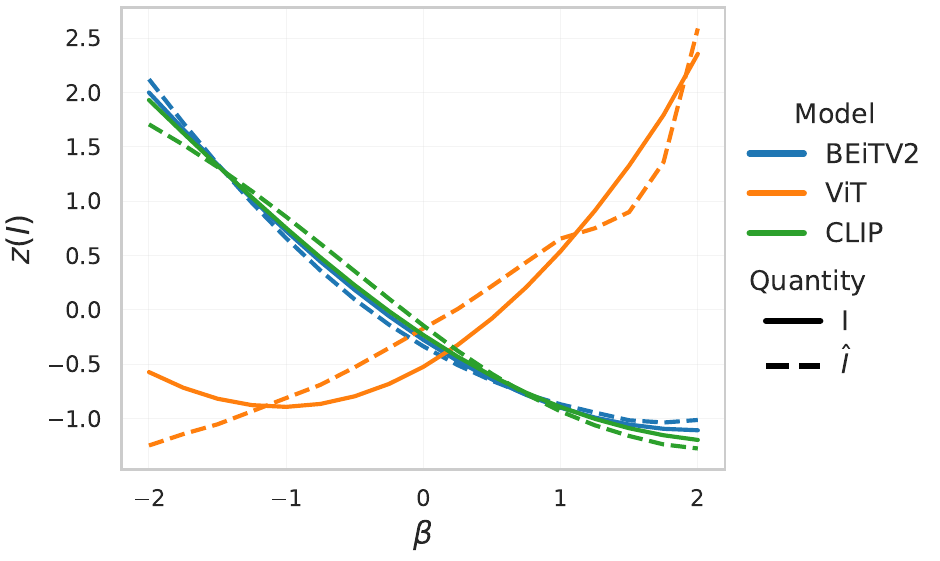}
  \caption{
  \textbf{UST proxy tracks instability along $\beta$.}
  For representative models, we plot $\mathcal I(\beta)=\Var(\log S_\beta)$ (solid) and the ID-only proxy $\widehat{\mathcal I}(\beta)$ (dashed),
  both standardized within each model across $\beta$.
  The proxy tracks the shape of the instability trajectory and predicts its minimizer $\hat\beta$ using ID-only quantities.
  (Summary statistics across all models appear in Appendix~\ref{app:ust_proxy}.)
  }
  \label{fig:ust_beta_overlay}
  \vspace{-0.8em}
\end{figure}

Motivated by Eq.~\eqref{eq:I_decomp}--\eqref{eq:W_eig}, we test whether \(\mathcal I(\beta)\) can be summarized by a small set of ID geometry statistics.
Within each model, we find that the instability curve is well captured by
\begin{equation}
\widehat{\mathcal I}(\beta)=a\,\log m_k(\beta)+b\,|s(\beta)|+c\,\mathrm{curv}(\beta),
\label{eq:Ihat}
\end{equation}
where \(m_k(\beta)\) is the $k$NN LID estimate and \((|s(\beta)|,\mathrm{curv}(\beta))\) summarizes the within-class spectrum
(Appendix~\ref{app:ust_defs}).
Figure~\ref{fig:ust_beta_overlay} shows representative overlays; aggregate correlations and fitting details are reported in Appendix~\ref{app:ust_proxy}.
Our practical \(\beta\)-selection in Sec.~\ref{sec:beta_proxy_selection} remains coefficient-free via \(P(\beta)=m(\beta)|s(\beta)|\). Notably, the $\beta$-instability trend is model-specific: in Fig.~\ref{fig:ust_beta_overlay}, $\mathcal I(\beta)$ increases with $\beta$ for ViT, whereas it decreases for BEiTV2 and CLIP. These opposing trajectories explain why the preferred $\beta$ can lie in different regions across models and motivate treating distinct curve shapes as separate regimes tied to $\mathcal I(\beta)$ and $\widehat{\mathcal I}(\beta)$.

\paragraph{Connection to $m(\beta)\,|s(\beta)|$.}
Eq.~\eqref{eq:Ihat} clarifies why $m(\beta)$ and $|s(\beta)|$ act as complementary coordinates:
$|s|$ summarizes spectral heterogeneity that amplifies point-dependent exposure in Eq.~\eqref{eq:W_eig},
while $m_k$ captures an additional local geometry signal that improves the prediction of $A_W(\beta)$ beyond spectral statistics alone
(see the coupling analysis in Appendix~\ref{app:ust_proxy}).
This provides a principled rationale for the empirical effectiveness of the simpler product $m(\beta)\,|s(\beta)|$ as a compact summary, while making explicit that it is necessarily lossy.

\section{Conclusion}
We studied Mahalanobis-style OOD detection across a range of vision foundation-model backbones, OOD benchmarks, and feature normalizations, and found that performance depends strongly on the representation. Across models and detector variants, two ID geometry signals, local intrinsic dimensionality and the within-class spectral slope, track this variation. We also introduced radially scaled $\ell_2$ normalization, which adjusts feature radii while keeping directions fixed, and an ID-only rule for choosing $\beta$. This selection improves over fixed normalizations in most settings and can approach oracle-tuned performance without using OOD samples.

\section*{Impact Statement}
This paper studies how representation geometry and feature normalization affect Mahalanobis-style OOD detection, with the goal of improving the reliability of deployed vision models. The primary positive impact is practical: our findings provide diagnostics and simple post-hoc normalization procedures that can reduce false positives and improve robustness monitoring in safety-relevant settings (e.g., medical imaging, autonomous systems, industrial inspection). Potential risks include misuse for overconfidence: improved OOD detection may be treated as a guarantee of safety, despite the fact that OOD detection remains imperfect and depends on data, model, and deployment conditions. In addition, geometry-based tuning could be adapted to evade certain detectors if an adversary can influence feature distributions. We therefore emphasize that our methods should be used as one component of a broader reliability pipeline (e.g., calibration, auditing, and monitoring), and that deployment decisions should not rely on OOD scores alone.

\bibliography{references}
\bibliographystyle{icml2026}

\newpage
\appendix
\onecolumn

\section{Detailed Description of Spectral and Manifold Metrics}
\label{app:metrics}

All spectral metrics are computed from the ordered eigenvalues $\lambda_1\ge\cdots\ge\lambda_d$ of a chosen matrix $M\in\{C,S_w,S_b\}$
(see Appendix~\ref{app:matrix_computation} for definitions). When needed, we use the normalized spectrum
$p_i \triangleq \lambda_i / \sum_{j=1}^d \lambda_j$.

\paragraph{Intrinsic dimensionality (ID).}
A global estimate of manifold dimension using the maximum-likelihood estimator of \citet{ma_characterizing_2018}.

\paragraph{Local intrinsic dimensionality (LID).}
For a feature vector $z$, let $r_j(z)$ denote the distance to its $j$-th nearest neighbor in feature space.
The $k$NN LID estimator is
\begin{equation}
\mathrm{LID}_k(z)
\;=\;
-\left[\frac{1}{k}\sum_{j=1}^{k}\log\frac{r_j(z)}{r_k(z)}\right]^{-1}.
\label{eq:lid_def_app}
\end{equation}
We report the dataset mean for $k\in\{10,25,50,100\}$.

\paragraph{Total variance (trace).}
\begin{equation}
\mathrm{Tr}(M) \;=\; \sum_{i=1}^d \lambda_i .
\label{eq:trace_app}
\end{equation}

\paragraph{Effective rank (trace-to-top ratio).}
We use the ratio
\begin{equation}
r_{\mathrm{eff}} \;\triangleq\; \frac{\sum_{i=1}^d \lambda_i}{\lambda_1}.
\label{eq:eff_rank_app}
\end{equation}

\paragraph{Participation ratio (PR).}
\begin{equation}
\mathrm{PR} \;\triangleq\; \frac{\left(\sum_{i=1}^d \lambda_i\right)^2}{\sum_{i=1}^d \lambda_i^2}.
\label{eq:pr_app}
\end{equation}

\paragraph{Condition number.}
\begin{equation}
\kappa \;\triangleq\; \frac{\lambda_1}{\lambda_d},
\label{eq:cond_app}
\end{equation}
where $\lambda_d$ is the smallest (non-zero) eigenvalue (or the smallest eigenvalue after numerical regularization, when applicable).

\paragraph{Spectral gap (head).}
\begin{equation}
\mathrm{Gap} \;\triangleq\; \lambda_1 - \lambda_6.
\label{eq:gap_app}
\end{equation}

\paragraph{Spectral entropy.}
\begin{equation}
\mathrm{H} \;\triangleq\; -\sum_{i=1}^d p_i \log p_i,
\qquad
p_i=\frac{\lambda_i}{\sum_j \lambda_j}.
\label{eq:entropy_app}
\end{equation}

\paragraph{Average log decay rate (top-20).}
\begin{equation}
\frac{1}{19}\sum_{i=1}^{19}\bigl(\log \lambda_i - \log \lambda_{i+1}\bigr).
\label{eq:avg_log_decay_app}
\end{equation}

\paragraph{Log-spectrum slope (full-range).}
We fit a least-squares line to the log-spectrum
\begin{equation}
\log \lambda_i \;=\; a + b\, i,
\label{eq:slope_fit_app}
\end{equation}
and report the fitted slope $b$ (in the main paper we often use $|s|$ for the magnitude of this slope when $s<0$).

\paragraph{Power-law exponent.}
We fit $\lambda_i \propto i^{-\beta}$ by regressing $\log\lambda_i$ on $\log i$ and report the exponent $\beta$.

\paragraph{Dimension for 90\% explained variance.}
\begin{equation}
k_{0.9}
\;\triangleq\;
\min\left\{k:\frac{\sum_{i=1}^{k}\lambda_i}{\sum_{j=1}^{d}\lambda_j}\ge 0.9\right\}.
\label{eq:dim90_app}
\end{equation}

These definitions match the metrics used in Section~\ref{sec:ideal_geometry} and enable exact reproducibility.

\section{Computation and Intuition for Covariance and Scatter Matrices}
\label{app:matrix_computation}

Our spectral analyses are based on three symmetric positive semidefinite matrices computed from in-distribution (ID) feature embeddings.
Let $z_i \in \mathbb{R}^d$ be the feature of sample $x_i$ with label $y_i\in\{1,\dots,K\}$, and let $N$ be the number of ID samples.
Define the global mean $\mu \triangleq \tfrac{1}{N}\sum_{i=1}^N z_i$, class means $\mu_k \triangleq \tfrac{1}{n_k}\sum_{i:y_i=k} z_i$, and class counts $n_k$.
All eigenvalues are reported in descending order, $\lambda_1\ge\cdots\ge\lambda_d$.

\paragraph{Global covariance ($C$).}
We measure the overall spread of ID features with
\begin{equation}
C \;\triangleq\; \frac{1}{N}\sum_{i=1}^{N} (z_i-\mu)(z_i-\mu)^\top .
\label{eq:cov_C_app}
\end{equation}
Large eigenvalues of $C$ correspond to directions of high variance across the entire ID dataset.

\paragraph{Within-class scatter ($S_w$).}
To measure intra-class variability we use the tied within-class scatter
\begin{equation}
S_w \;\triangleq\; \frac{1}{N}\sum_{k=1}^{K}\;\sum_{i:y_i=k} (z_i-\mu_k)(z_i-\mu_k)^\top .
\label{eq:Sw_app}
\end{equation}
When the Mahalanobis detector is implemented with a shared (tied) covariance across classes, its covariance estimate coincides with $S_w$.
In particular, the standard class-conditional MD score can be written as
$\mathrm{MD}_k(z)=(z-\mu_k)^\top S_w^{-1}(z-\mu_k)$, matching Eq.~\ref{eq:md-eigen} in the main text.

\paragraph{Between-class scatter ($S_b$).}
To quantify how class means spread around the global mean, we use
\begin{equation}
S_b \;\triangleq\; \frac{1}{N}\sum_{k=1}^{K} n_k\,(\mu_k-\mu)(\mu_k-\mu)^\top .
\label{eq:Sb_app}
\end{equation}
Large eigenvalues of $S_b$ indicate directions along which class centroids are well separated.

\paragraph{Spectral Shift Metric.}
To study how representations change under distributional shifts, for each matrix $M \in \{C, S_w, S_b\}$ we compute its eigenvalues
$\{\lambda_i^{\text{train}}\}$ on the training set
and $\{\lambda_i^{\text{eval}}\}$ on a validation or OOD set.
The relative eigenvalue shift is defined as
\begin{equation}
\Delta_i(M)
  = \frac{\lambda_i^{\text{eval}} - \lambda_i^{\text{train}}}
         {\lambda_i^{\text{train}}}.
\end{equation}
This spectrum of shifts highlights how the geometry of the representation
changes under distributional shift, providing a fine-grained indicator of
robustness or overfitting.

\paragraph{Intuition Behind the Shift Metric.}  
\begin{itemize}
    \item \textbf{Zero shift ($\Delta_i \approx 0$):} The corresponding direction in feature space is stable across data splits.  
    \item \textbf{Positive shift ($\Delta_i > 0$):} The representation spreads out along this eigenvector in the new data, increasing variance.  
    \item \textbf{Negative shift ($\Delta_i < 0$):} The representation compresses along this eigenvector, reducing variance.  
    \item \textbf{Magnitude:} Reflects the relative degree of expansion or contraction. For example, $\Delta_i = 0.5$ indicates a 50\% increase in variance, while $\Delta_i = -0.2$ indicates a 20\% decrease.
\end{itemize}

\paragraph{Interpretation in Model Analysis.}  
\begin{itemize}
    \item \textbf{Small shifts across all eigenvectors:} Robust and stable representations that generalize well.  
    \item \textbf{Large positive shifts:} Features become more variable on new data, potentially indicating under-regularization or sensitivity to OOD inputs.  
    \item \textbf{Large negative shifts:} Features compress on new data, potentially indicating overfitting.  
    \item \textbf{Consistent shift patterns:} Systematic changes in representation geometry, revealing overfitting or robustness issues.
\end{itemize}

\paragraph{Types of Shifts.}  
\begin{itemize}
    \item \textbf{Validation covariance shift:} Change in global covariance from training to validation data.  
    \item \textbf{OOD covariance shift:} Change in global covariance from training to out-of-distribution data.  
    \item \textbf{Validation within-class shift:} Change in within-class scatter from training to validation data.  
    \item \textbf{Validation between-class shift:} Change in between-class scatter from training to validation data.  
\end{itemize}

\section{Spectral analysis of training effects}
\label{app:spectral_analysis}
To understand how the intrinsic geometry of representations affects OOD performance, we begin by examining the spectral properties of three key matrices: the feature covariance $C$, the within-class scatter $S_w$, and the between-class scatter $S_b$. These matrices capture complementary aspects of the feature space: $C$ reflects overall variance, $S_w$ measures intra-class dispersion, and $S_b$ quantifies inter-class separation (more details in Appendix~\ref{app:matrix_computation}).  
Our first analysis focuses on the eigenvalue spectra of these matrices. The magnitude and decay of eigenvalues reveal how variance is distributed across dimensions, providing insight into the richness and anisotropy of the feature space. For instance, a steep decay in $S_w$ eigenvalues indicates that intra-class variability is concentrated along a few directions, resulting in tight clusters, whereas a slower decay suggests more diffuse intra-class variation. Similarly, large eigenvalues in $S_b$ correspond to well-separated class means, signaling strong discriminability.

\begin{figure}[h!]
    \centering
    \includegraphics[width=0.8\linewidth]{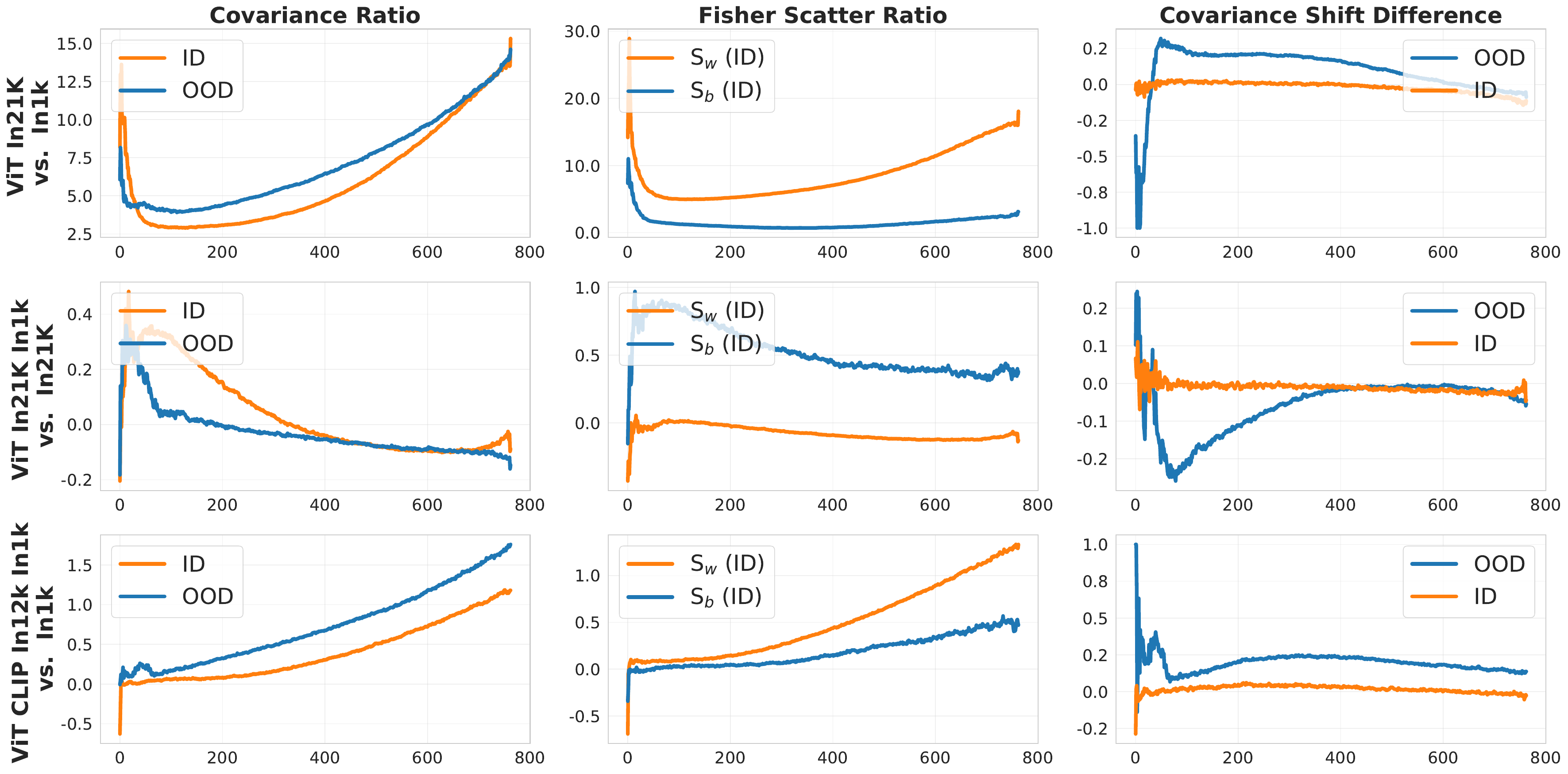}
    \caption{Spectral ratios across models. Higher ratios indicate richer within-class variation and more expressive feature spaces. Fine-tuning tends to increase $S_b$ while preserving $S_w$.}
    \label{fig:eigvals_diff_by_model}
    \vspace{-1.0em}
\end{figure}

\paragraph{Spectral ratios}  
To systematically compare models, we compute ratios between eigenvalues of $S_b$, $S_w$, and $C$. These ratios serve as compact summaries of representation geometry. Higher $S_b / S_w$ ratios indicate representations with greater between-class separation relative to intra-class spread, which generally favors OOD detection, while lower ratios may signal overlapping clusters or limited discriminative power. A higher $C$ ratio indicates that variance is distributed along multiple directions, reflecting a richer and more expressive representation that can better accommodate novel OOD inputs without major distortion. As illustrated in Figure \ref{fig:eigvals_diff_by_model}, models pretrained on large, diverse datasets (e.g., In21k) exhibit larger $C$ and $S_w$ ratios, capturing richer intra-class variations and producing more expressive feature spaces. Fine-tuning tends to increase $S_b$ ratios while preserving $S_w$, enhancing class separability without sacrificing cluster compactness. Models trained on smaller datasets exhibit smaller ratios, reflecting less expressive representations with weaker discriminability.

\paragraph{Eigenvalue shifts}  
Beyond static spectra, we are interested in how stable the representation geometry is under distributional shifts. To capture this, we define a \emph{spectral shift metric}, which measures the relative change in eigenvalues from the training set to validation or OOD data (see Appendix~\ref{app:matrix_computation}). A small shift indicates that the representation preserves its structure across data splits, signaling robustness. Large positive shifts reveal that features are spreading along new directions, while large negative shifts indicate compression. Figure \ref{fig:eigvals_diff_by_model} shows that OOD samples induce larger spectral shifts in models trained on small datasets, reflecting lower generalization and brittle feature structures. Large-scale pretrained models show smaller shifts, indicating more stable, robust representations under distributional change. Fine-tuning generally maintains small shifts while increasing $S_b$, improving class separation without compromising intra-class compactness.

\section{Geometry Induced by Radially Scaled $\ell_2$ Normalization and $\beta$ Selection}
\label{app:radial-metric}

This appendix provides geometric intuition for the radially scaled $\ell_2$ map $\phi_\beta$ (Sec.~\ref{sec:radial_l2})
and specifies the ID-only $\beta$ selection rule used throughout the paper (Sec.~\ref{sec:beta_proxy_selection}).
Our goal is modest: (i) show that $\beta$ continuously reweights \emph{radial} versus \emph{angular} variations, explaining
why both LID $m(\beta)$ and within-class spectral summaries $s(\beta)$ vary smoothly with $\beta$; and (ii) formalize how
we choose $\hat\beta$ from the proxy curve $P(\beta)$ without committing a priori to ``minimize'' or ``maximize.''

\subsection{Setup and induced metric}
Let $z\in\mathbb{R}^d\setminus\{0\}$ be a feature vector with Euclidean metric $g_{\mathrm{Euc}}$.
Consider the radial map
\begin{equation}
\phi_\beta(z)\;=\;\frac{z}{\|z\|^\beta},
\qquad \beta\in\mathbb{R},
\label{eq:phi_beta_app}
\end{equation}
which preserves direction but rescales radius.
Writing $z=r\,u$ with $r=\|z\|$ and $u\in S^{d-1}$, the Euclidean metric decomposes as
\begin{equation}
g_{\mathrm{Euc}} \;=\; \mathrm{d}r^2 + r^2 g_{S^{d-1}}.
\label{eq:euc_polar_app}
\end{equation}
Under $\phi_\beta$, the radius becomes $R=r^{1-\beta}$ and $\mathrm{d}R=(1-\beta)\,r^{-\beta}\mathrm{d}r$.
The pullback metric $g_\beta \triangleq \phi_\beta^\ast g_{\mathrm{Euc}}$ therefore satisfies
\begin{equation}
g_\beta
\;=\;
(1-\beta)^2 r^{-2\beta}\,\mathrm{d}r^2
\;+\;
r^{2(1-\beta)} g_{S^{d-1}} .
\label{eq:radial_metric}
\end{equation}
Eq.~\eqref{eq:radial_metric} makes explicit that $\beta$ changes the relative weighting of radial and angular variations.
This is the geometric reason that empirical quantities computed \emph{after} $\phi_\beta$
(e.g., covariance spectra and neighborhood distances used for LID) vary smoothly with $\beta$.

\subsection{How this interacts with Mahalanobis-style scoring}
In the main paper we apply $\phi_\beta$ to features and then fit the (tied) Gaussian statistics used by Mahalanobis variants.
Equivalently, $\phi_\beta$ changes the distribution of deviations $\delta_\beta(z)$ before inserting them into a quadratic score
$S_\beta(z)=\delta_\beta(z)^\top \Sigma(\beta)^{-1}\delta_\beta(z)$.
Thus, $\beta$ does not define a new detector; it defines a one-parameter family of \emph{geometrically deformed representations}.
Through Eq.~\eqref{eq:radial_metric}, this deformation changes:
(i) neighborhood structure (hence $m(\beta)$), (ii) scatter spectra (hence $s(\beta)$), and (iii) how quadratic weighting
amplifies directional deviations.

\subsection{Interpreting $\beta$}
Eq.~\eqref{eq:radial_metric} yields a compact interpretation of typical regimes:
\begin{itemize}[leftmargin=1.2em, itemsep=0.2em]
    \item \textbf{$\beta=0$ (identity).} $\phi_\beta$ is the identity and $g_\beta=g_{\mathrm{Euc}}$.
    \item \textbf{$0<\beta<1$ (moderate contraction).} Radii shrink as $r^{1-\beta}$ while angular structure is retained.
    \item \textbf{$\beta=1$ (spherical projection).} Radii become constant ($R=1$): radial variability is removed and only angles remain.
    \item \textbf{$\beta>1$ (strong contraction).} Large radii are aggressively compressed; this can suppress variability but may also collapse useful structure.
    \item \textbf{$\beta<0$ (expansion).} Radii are amplified; norm differences become more prominent, which can increase sensitivity to radial outliers.
\end{itemize}

\subsection{$\beta$ selection from the proxy curve}
\label{app:beta_selection_rule}

For each model we evaluate the proxy
\begin{equation}
P(\beta)=m(\beta)\,|s(\beta)|
\label{eq:proxy_app}
\end{equation}
on a discrete grid $\mathcal B=\{\beta_1<\cdots<\beta_T\}$.
Empirically, $P(\beta)$ is usually inverted-U shaped (so maximizing $P$ is the typical behavior), but some
representations can produce U-shaped curves (where minimizing $P$ is appropriate). Rather than hard-coding
``maximize'' versus ``minimize,'' we select the most pronounced \emph{interior} turning point.

\paragraph{Interior turning point rule.}
Let $P_t \triangleq P(\beta_t)$. We first define a simple endpoint reference level
\[
P_{\mathrm{end}} \;\triangleq\; \frac{P_1+P_T}{2}.
\]
Among interior grid points $t\in\{2,\ldots,T-1\}$, we choose the one farthest from this endpoint level:
\begin{equation}
\hat t \in \arg\max_{t\in\{2,\ldots,T-1\}} \big|P_t - P_{\mathrm{end}}\big|,
\qquad
\hat\beta \triangleq \beta_{\hat t}.
\label{eq:beta_rule}
\end{equation}
This rule returns the interior \emph{maximum} when the curve is inverted-U, and the interior \emph{minimum}
when the curve is U-shaped, while discouraging endpoint-driven choices.

\paragraph{Nearly monotone curves.}
If the proxy has no clear interior turning point on the finite grid (i.e., the largest interior deviation is very small),
we fall back to the default behavior and set
\[
\hat\beta \in \arg\max_{\beta\in\mathcal B} P(\beta).
\]
In our experiments, this situation is rare and typically occurs when the true extremum lies outside the evaluated range.

\paragraph{Connection to the main text.}
The selection rule is intentionally agnostic to whether $P(\beta)$ should be maximized or minimized.
In Sec.~\ref{sec:ust_factorization} we further explain why both U-shaped and inverted-U regimes occur by relating
the proxy curve to the instability functionals $\mathcal I(\beta)$ and its low-dimensional approximation
$\widehat{\mathcal I}(\beta)$.

\section{Unified Stability Theory: Definitions, Mechanism, and Empirical Evidence}
\label{app:ust}

This appendix provides the formal definitions used by Unified Stability Theory (UST), derives the stretch mechanism
underlying the main text, and summarizes the supporting empirical evidence. Unless stated otherwise, all quantities are
computed on last-layer features and estimated using ID data only.

\subsection{Quadratic scores and the instability functional}
\label{app:ust_defs}
Each Mahalanobis-style detector specifies a centered deviation $\delta_\beta(z)$, where $z$ is a feature vector, and a tied scatter (or covariance)
estimate $\Sigma(\beta)$ computed from ID data. 
We assume $\Sigma(\beta)\succ 0$ for all $\beta$ under consideration, ensuring $\Sigma(\beta)^{-1}$ exists, and that $\|\delta_\beta(z)\|>0$ almost surely under $\mathcal D_{\mathrm{ID}}$ so $\log\|\delta_\beta(z)\|^2$ is well-defined.

The corresponding quadratic score is
\begin{equation}
S_\beta(z)\;\triangleq\;\delta_\beta(z)^\top \Sigma(\beta)^{-1}\delta_\beta(z).
\label{eq:ust_score_app}
\end{equation}
We quantify ID-side sensitivity of the score using
\begin{equation}
\mathcal I(\beta)\;\triangleq\;\operatorname{Var}_{z\sim\mathcal D_{\mathrm{ID}}}\!\big[\log S_\beta(z)\big].
\label{eq:ust_I_app}
\end{equation}

\begin{figure}[h]
  \centering
  \includegraphics[width=0.7\linewidth]{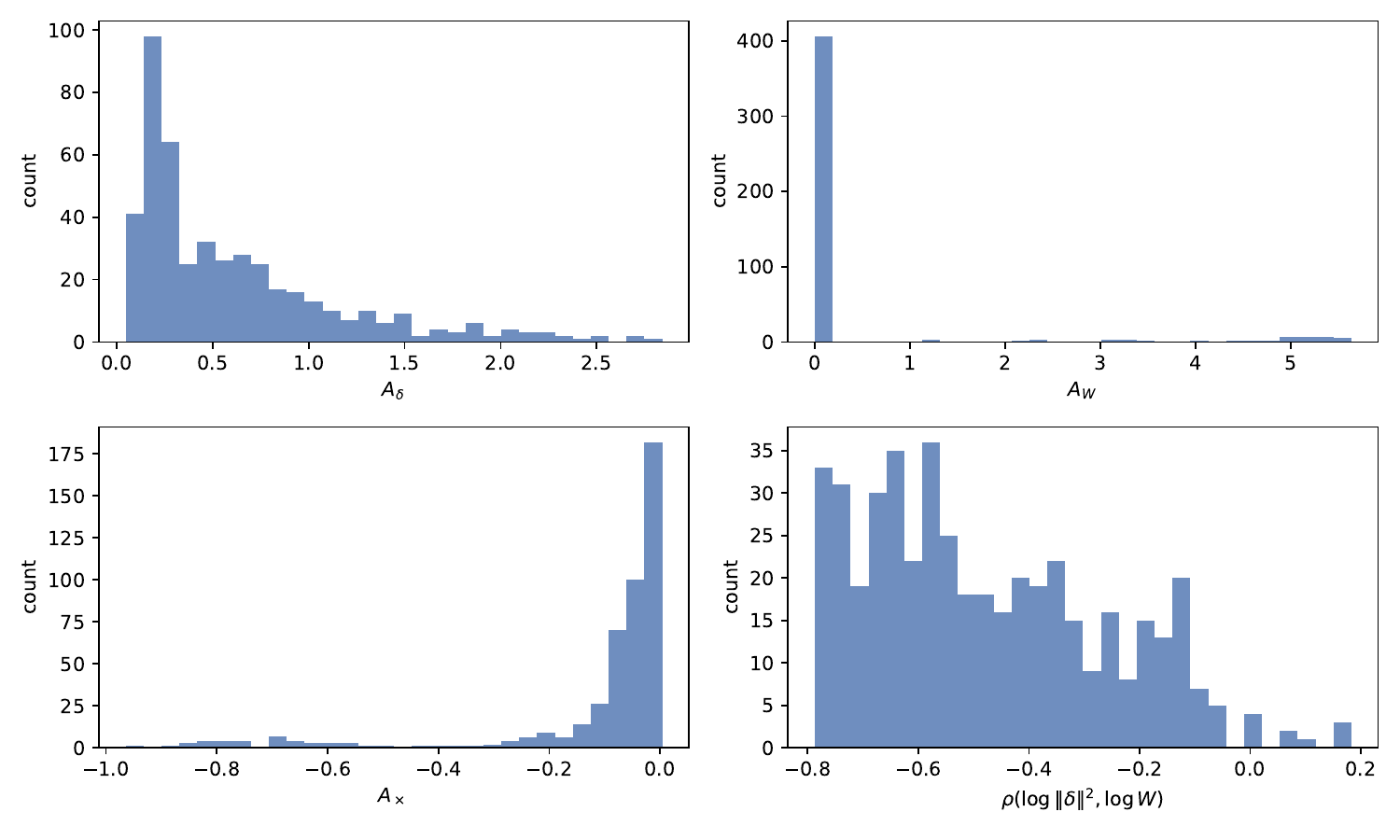}
  \caption{Distribution of the three terms $(A_\delta,A_W,A_\times)$ across models and $\beta$.
  The cross-term is predominantly negative, consistent with compensation between size and stretch.}
  \label{fig:ust_channel_summary}
\end{figure}

\paragraph{Exact factorization and channel decomposition.}
For $\delta_\beta(z)\neq 0$, define
\begin{equation}
W_\beta(z)\;\triangleq\;\frac{\delta_\beta(z)^\top\Sigma(\beta)^{-1}\delta_\beta(z)}{\|\delta_\beta(z)\|^2}.
\label{eq:ust_W_app}
\end{equation}
Then
\begin{equation}
S_\beta(z)=\|\delta_\beta(z)\|^2\,W_\beta(z),
\qquad
\log S_\beta(z)=\log\|\delta_\beta(z)\|^2+\log W_\beta(z).
\label{eq:ust_factorization_app}
\end{equation}
We verify the factorization in Eq.~\eqref{eq:ust_factorization_app} numerically to machine precision (Table~\ref{tab:ust_factorization_residuals}). 
Taking the variance over ID samples yields the exact decomposition
\begin{equation}
\mathcal I(\beta)=A_\delta(\beta)+A_W(\beta)+2A_\times(\beta),
\label{eq:ust_channel_decomp_app}
\end{equation}
where
\begin{equation}
\begin{aligned}
A_\delta(\beta) &\triangleq \operatorname{Var}_{z\sim\mathcal{D}_{\mathrm{ID}}}\!\big[\log\|\delta_\beta(z)\|^2\big],\\
A_W(\beta) &\triangleq \operatorname{Var}_{z\sim\mathcal{D}_{\mathrm{ID}}}\!\big[\log W_\beta(z)\big],\\
A_\times(\beta) &\triangleq \operatorname{Cov}_{z\sim\mathcal{D}_{\mathrm{ID}}}\!\big(\log\|\delta_\beta(z)\|^2,\log W_\beta(z)\big).
\end{aligned}
\label{eq:ust_channel_defs_app}
\end{equation}
In the main text we refer to $A_\delta$ as the \emph{size} channel and $A_W$ as the \emph{stretch} channel. Across models and $\beta$, both channels $A_\delta(\beta)$ and $A_W(\beta)$ exhibit nontrivial variation, and the cross-term $A_\times(\beta)$ is predominantly negative (Fig.~\ref{fig:ust_channel_summary}), indicating partial compensation between size and stretch.

\paragraph{Empirical evidence.}
We numerically verify Eq.~\eqref{eq:ust_factorization_app} to machine precision (Table~\ref{tab:ust_factorization_residuals}),
and we observe substantial variation in both $A_\delta(\beta)$ and $A_W(\beta)$ across $\beta$ and models
(Figure~\ref{fig:ust_channel_summary}). Moreover, the cross-term is systematically negative in our configurations,
indicating partial compensation between size and stretch (also summarized in Figure~\ref{fig:ust_channel_summary}).

\begin{table}[t]
\centering
\caption{Numerical residual for the factorization $S_\beta(z)=\|\delta_\beta(z)\|^2W_\beta(z)$ over all evaluated configurations.}
\label{tab:ust_factorization_residuals}
\begin{tabular}{lccc}
\toprule
Statistic & max rel.~err & mean rel.~err & \#configs \\
\midrule
All & 7.34e-07 & 4.56e-08 & 462 \\
\bottomrule
\end{tabular}
\end{table}

\subsection{Stretch mechanism: spectrum \texorpdfstring{$\times$}{x} allocation geometry}
\label{app:ust_stretch}
This subsection derives the representation of $W_\beta$ that motivates the intervention experiments.

Write the eigendecomposition $\Sigma(\beta)=U(\beta)\Lambda(\beta)U(\beta)^\top$, where $\Lambda(\beta)=\mathrm{diag}(\lambda_1(\beta),\ldots,\lambda_d(\beta))$ with $\lambda_1(\beta)\geq\cdots\geq\lambda_d(\beta)>0$, and $U(\beta)=[u_1(\beta),\ldots,u_d(\beta)]$ is an orthonormal eigenbasis.

Define the allocation of a deviation $\delta_\beta(z)$ onto eigen-directions by
\begin{equation}
p_i(z;\beta)\triangleq\frac{(u_i(\beta)^\top\delta_\beta(z))^2}{\|\delta_\beta(z)\|^2},
\qquad \sum_{i=1}^d p_i(z;\beta)=1.
\label{eq:ust_p_def_app}
\end{equation}
When $\beta$ is clear from context we abbreviate $p_i(z;\beta)$ as $p_i(z)$.
Substituting the eigendecomposition into Eq.~\eqref{eq:ust_W_app} gives the exact identity
\begin{equation}
W_\beta(z)=\sum_{i=1}^d \frac{p_i(z)}{\lambda_i(\beta)}.
\label{eq:ust_W_eig_app}
\end{equation}

\paragraph{Interpretation.}
Equation~\eqref{eq:ust_W_eig_app} shows that pointwise exposure is controlled jointly by
(i) \emph{spectral heterogeneity} through $\{\lambda_i(\beta)\}$ and
(ii) \emph{allocation geometry} through the random simplex vector $p(z)$.
In particular, variability in $W_\beta$ (and hence $A_W$) can be reduced either by flattening the spectrum
(e.g., shrinkage, spectral shaping) or by making allocations more uniform across samples
(e.g., reducing dominance/spikiness or explicitly smoothing $p(z)$).

\subsection{Interventions isolating size and stretch}
\label{app:ust_interventions}

We now summarize the intervention experiments that isolate the two channels in Eq.~\eqref{eq:ust_channel_decomp_app}.
Each intervention targets a single mechanism suggested by Eq.~\eqref{eq:ust_W_eig_app}.

\begin{figure}[t]
  \centering
  \includegraphics[width=0.5\linewidth]{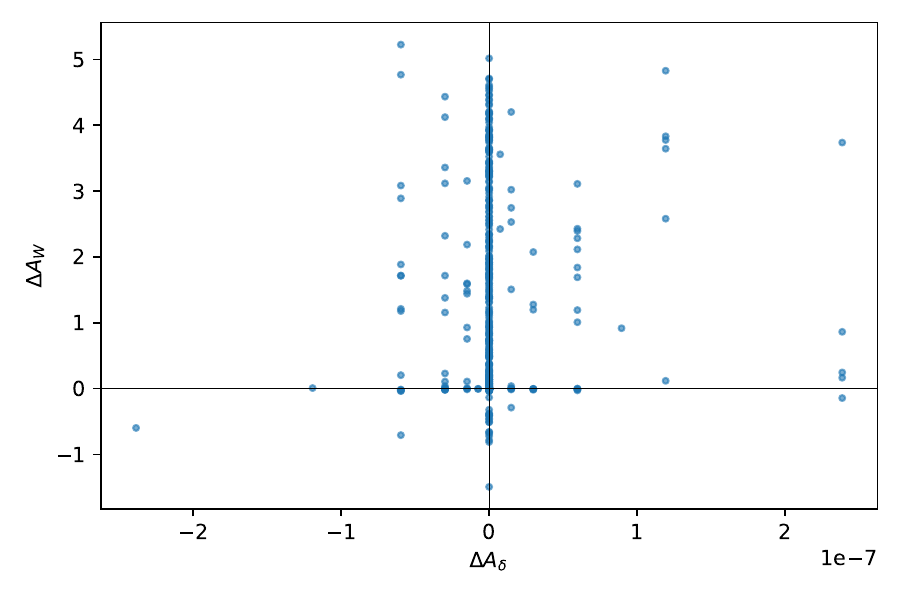}
  \caption{Rotation intervention: paired changes $(\Delta A_\delta,\Delta A_W)$ under orthogonal mixing. $\Delta A_\delta$ concentrates near zero (numerical scale $\sim10^{-7}$), whereas $\Delta A_W$ varies substantially, indicating that rotation primarily perturbs stretch variability.}
  \label{fig:ust_rotation}
\end{figure}

\paragraph{Rotation changes allocation geometry while preserving norms.}
Applying a random orthogonal map to deviations, $\delta_\beta(z)\mapsto Q\,\delta_\beta(z)$ with $Q^\top Q=I$,
leaves $\|\delta_\beta(z)\|$ unchanged (up to numerical precision) but alters the projections
$\{u_i(\beta)^\top \delta_\beta(z)\}_{i=1}^d$. Hence, the allocation vector $p(z)$ in Eq.~\eqref{eq:ust_W_eig_app}
is redistributed across eigen-directions, which directly perturbs the exposure term $W_\beta(z)$.
Empirically, the rotation intervention induces large changes in $A_W$ while $\Delta A_\delta$ stays near zero across configurations (Fig.~\ref{fig:ust_rotation}).

\begin{figure}[t]
  \centering
  \includegraphics[width=0.6\linewidth]{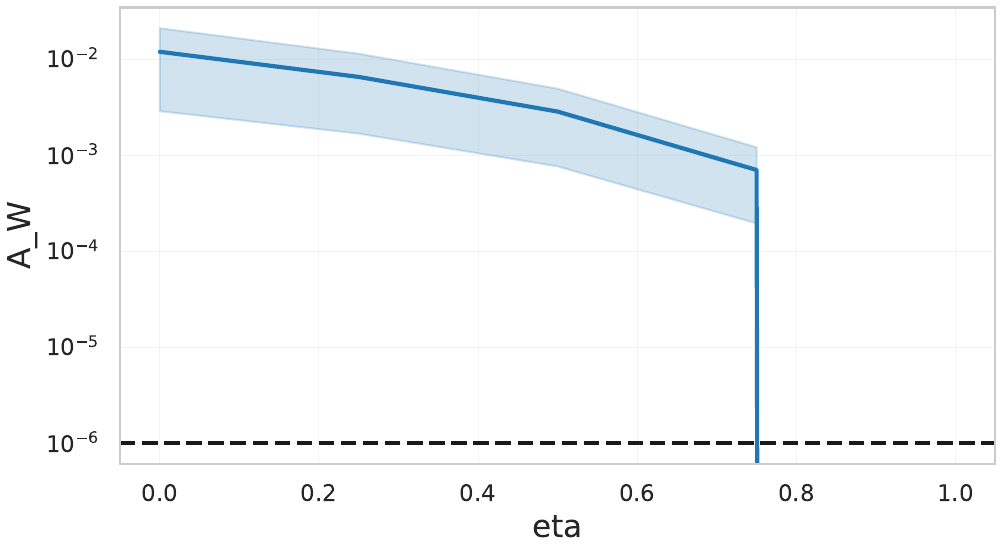}
  \caption{Allocation smoothing: $A_W(\eta)$ decreases and collapses as $\eta\to 1$, consistent with Eq.~\eqref{eq:ust_W_eig_app}.}
  \label{fig:ust_allocation_smoothing}
\end{figure}

\paragraph{Allocation smoothing collapses stretch variability.}
We construct a controlled family of allocations
\begin{equation}
p^{(\eta)}(z)=(1-\eta)p(z)+\eta\,\bar p,\qquad \eta\in[0,1],
\end{equation}
where $\bar p$ is the ID-average allocation. Substituting $p^{(\eta)}$ into Eq.~\eqref{eq:ust_W_eig_app}
removes point-dependent exposure as $\eta\to 1$.
Empirically, $A_W(\eta)$ decreases monotonically and approaches numerical zero at $\eta=1$
(Figure~\ref{fig:ust_allocation_smoothing}).

\paragraph{Spectrum shaping changes $A_W$ with allocations held fixed.}
We modify the eigenvalues while keeping the eigenvectors fixed (e.g., $\lambda_i'\propto \lambda_i^\alpha$ with trace renormalization) 
and recompute $W_\beta$ using the same allocations $p(z)$ in Eq.~\eqref{eq:ust_W_eig_app}.
This directly tests the spectral factor in the exposure mechanism and yields large changes in $A_W$.

\paragraph{Radial rescaling leaves \(W\) invariant under fixed \(\Sigma\).}
If \(\Sigma(\beta)\) is fixed and deviations are rescaled as \(\delta'(z)=c(z)\delta(z)\) for any \(c(z)>0\), then
\[
W'(z)
=\frac{\delta'(z)^\top\Sigma(\beta)^{-1}\delta'(z)}{\|\delta'(z)\|^2}
=\frac{c(z)^2\,\delta(z)^\top\Sigma(\beta)^{-1}\delta(z)}{c(z)^2\,\|\delta(z)\|^2}
=W(z).
\]
Therefore \(A_W=\Var(\log W)\) is invariant under such rescalings when \(\Sigma(\beta)\) is held fixed.
We verify this fixed-\(\Sigma\) invariance numerically in Table~\ref{tab:ust_scale_invariance_aw}.
For completeness, we also report a refit-\(\Sigma\) variant as a sensitivity analysis, where invariance is not expected. We include this invariance as a sanity check to ensure that our size-only manipulations cannot affect the exposure term $W$ unless the whitening operator $\Sigma^{-1}$ itself changes.

\begin{table}[h]
\centering
\caption{Scale invariance of \(A_W\) under fixed \(\Sigma\).}
\label{tab:ust_scale_invariance_aw}
\begin{tabular}{lcc}
\toprule
Metric & mean \(|\Delta A_W|\) & \#configs \\
\midrule
E8 & $<10^{-6}$ & 1386 \\
\bottomrule
\end{tabular}
\end{table}

\subsection{Low-dimensional proxy and the role of \texorpdfstring{$m_k$}{mk}}
\label{app:ust_proxy}

This subsection supports the claim in Sec.~\ref{sec:ust_proxy} that, within a fixed representation, the ID instability
trajectory $\beta\mapsto \mathcal I(\beta)$ can be captured by a small set of ID-only geometry statistics.
We consider the additive proxy
\begin{equation}
\widehat{\mathcal I}(\beta)
\;=\;
a\,\log m_k(\beta)\;+\;b\,|s(\beta)|\;+\;c\,\mathrm{curv}(\beta),
\label{eq:ust_I_hat_app}
\end{equation}
where $m_k(\beta)$ is the $k$NN LID estimate in $\phi_\beta$-space and $(|s(\beta)|,\mathrm{curv}(\beta))$ summarize the
eigenspectrum of the relevant scatter matrix (e.g., $\Sigma(\beta)$ or $S_w(\beta)$).

\paragraph{Spectral descriptors.}
Let $\lambda_1(\beta)\ge\cdots\ge\lambda_d(\beta)>0$ denote the ordered eigenvalues and define the log-spectrum
$\ell_i(\beta)\triangleq \log\lambda_i(\beta)$. We fit an affine model
\[
\ell_i(\beta)\;\approx\;\alpha(\beta)+s(\beta)\,i,\qquad i=1,\ldots,d,.
\]
by least squares and report:
\begin{itemize}
\item \textbf{Slope.} $s(\beta)$ is the fitted slope; we use $|s(\beta)|$ as a scale-free measure of overall spectral decay
(steeper decay $\Rightarrow$ larger $|s|$).
\item \textbf{Curvature (residual curvature).} We measure nonlinearity of the log-spectrum by the lack-of-fit of the same
linear regression:
\[
\mathrm{curv}(\beta)\;\triangleq\;1-R^2(\beta),
\]
so $\mathrm{curv}(\beta)\approx 0$ when $\log\lambda_i(\beta)$ is close to affine in $i$ (near-exponential eigen-decay), and
larger values indicate a more non-affine/irregular spectrum. Intuitively, this is the fraction of log-spectrum variance not explained by a straight line.
\end{itemize}

\paragraph{Per-model least squares (ID-only).}
Fix a representation (model) and evaluate the proxy components on a discrete grid
$\mathcal B=\{\beta_1,\ldots,\beta_T\}$ using ID features only. Define
\[
x(\beta_t)=\big[\log m_k(\beta_t),\ |s(\beta_t)|,\ \mathrm{curv}(\beta_t)\big]\in\mathbb{R}^3,
\qquad
y(\beta_t)=\mathcal I(\beta_t).
\]
We fit $(a,b,c)$ by least squares \emph{without} an intercept:
\[
(a,b,c)\ \in\ \arg\min_{a,b,c}\ \sum_{t=1}^T \Big(y(\beta_t)-\langle (a,b,c),\,x(\beta_t)\rangle\Big)^2.
\]
We omit an intercept since we evaluate agreement across $\beta$ (e.g., via rank correlation); an additive constant would not
change the ordering of $\widehat{\mathcal I}(\beta)$, and the proxy is intended to capture the \emph{shape} of the instability trajectory.

\paragraph{Proxy tracking along the $\beta$ trajectory.}
Within each model, $\widehat{\mathcal I}(\beta)$ tracks $\mathcal I(\beta)$ over the $\beta$ sweep with high rank agreement
(Table~\ref{tab:ust_proxy_tracking}; main-paper visualization: the overlay plot in Figure~\ref{fig:ust_beta_overlay}).

\paragraph{Coupling beyond spectrum.}
Regressing $A_W(\beta)$ on $(|s(\beta)|,\mathrm{curv}(\beta))$ and then adding $\log m_k(\beta)$ yields a consistent improvement
in explained variance across models (Median $\Delta R^2\ge 0.05$; mean $\Delta R^2$ is typically higher, with a wide IQR, indicating heterogeneous gains across model families).
This supports treating $m_k$ as an additional ID geometry statistic that modulates exposure-driven stretch variability.

\begin{table}[h]
\centering
\caption{Proxy tracking: within-model rank correlation between $\widehat{\mathcal I}(\beta)$ and $\mathcal I(\beta)$ over $\beta$.}
\label{tab:ust_proxy_tracking}
\begin{tabular}{lcc}
\toprule
$k$ & median Spearman $\rho$ & IQR \\
\midrule
10 & 0.532 & 0.670 \\
25 & 0.536 & 0.670 \\
50 & 0.540 & 0.670 \\
100 & 0.547 & 0.663 \\
\bottomrule
\end{tabular}
\end{table}

\subsection{RMD: Term-wise Unified Stability Lens}
\label{app:rmd_ust}

RMD is commonly implemented as a difference of two Mahalanobis-type quadratic forms,
\(
S_{\mathrm{RMD},\beta}(z)=S^{(1)}_\beta(z)-S^{(2)}_\beta(z)
\),
where the deviations \(\delta^{(j)}_\beta\) and scatter estimates \(\Sigma^{(j)}(\beta)\) may differ between terms.
As a result, RMD does not generally admit the single-\(\Sigma\) representation in Eq.~\eqref{eq:S_def}.

\subsection{Exact term-wise factorization and channels}
For each term \(j\in\{1,2\}\), define
\[
S^{(j)}_\beta(z)=\delta^{(j)}_\beta(z)^\top \Sigma^{(j)}(\beta)^{-1}\delta^{(j)}_\beta(z),
\qquad
W^{(j)}_\beta(z)=\frac{S^{(j)}_\beta(z)}{\|\delta^{(j)}_\beta(z)\|^2}.
\]
Then the exact factorization holds term-wise:
\[
S^{(j)}_\beta(z)=\|\delta^{(j)}_\beta(z)\|^2\,W^{(j)}_\beta(z),
\qquad
\log S^{(j)}_\beta(z)=\log\|\delta^{(j)}_\beta(z)\|^2+\log W^{(j)}_\beta(z),
\]
and the instability functional for each term decomposes as
\[
\mathcal I^{(j)}(\beta)\triangleq \Var[\log S^{(j)}_\beta]
=
A^{(j)}_\delta(\beta)+A^{(j)}_W(\beta)+2A^{(j)}_\times(\beta),
\]
with the obvious definitions of \(A^{(j)}_\delta,A^{(j)}_W,A^{(j)}_\times\).

\clearpage

\section{Full results}

\subsection{Cross-model performance}
\label{app:cross_model_performance}
This section reports the complete cross-model results corresponding to Figure~\ref{fig:ood_study_mds_ninco_compact}. Table~\ref{tab:cross_model_md_performance} provides detailed false positive rates (FPR@95) across all evaluated architectures and OOD datasets for both Mahalanobis distance (MD) and its regularized variant (RMD). These results complement the trends discussed in the main text: RMD typically outperforms MD across architectures and training regimes, with particularly pronounced gains for models pretrained on large-scale datasets without ImageNet-1k fine-tuning.

\begin{table}[ht]
\centering
\caption{False positive rate across different models and datasets using two Mahalanobis distance variants (MD and RMD)}
\adjustbox{max width=0.99\textwidth}{
\begin{tabular}{lrrrrrr@{\hskip 10mm}rrrrrr}
 \toprule
 & \multicolumn{6}{c}{Mahalanobis} & \multicolumn{6}{c}{Relative Mahalanobis} \\
\hline
Model & NINCO & OpO & SSB & Textures & iNat & Average & NINCO & OpO & SSB & Textures & iNat & Average \\
\hline
BEiTV2 In1k & 0.506 & 0.212 & 0.825 & 0.327 & 0.142 & 0.402 & 0.456 & 0.207 & 0.827 & 0.323 & 0.140 & \bfseries 0.391 \\
BEiTV2 In21k & 0.475 & 0.331 & 0.781 & 0.418 & 0.174 & 0.436 & 0.339 & 0.165 & 0.786 & 0.314 & 0.051 & \bfseries 0.331 \\
MAE In1k & 0.516 & 0.272 & 0.827 & 0.353 & 0.208 & 0.435 & 0.463 & 0.251 & 0.828 & 0.360 & 0.186 & \bfseries 0.418 \\
DINOV2 & 0.424 & 0.178 & 0.773 & 0.302 & 0.014 & \bfseries 0.338 & 0.577 & 0.200 & 0.845 & 0.415 & 0.032 & 0.414 \\
ViT & 0.557 & 0.302 & 0.843 & 0.376 & 0.206 & 0.457 & 0.516 & 0.303 & 0.806 & 0.417 & 0.206 & \bfseries0.450 \\
ViT In21K & 0.641 & 0.513 & 0.807 & 0.541 & 0.158 & 0.532 & 0.451 & 0.253 & 0.816 & 0.392 & 0.080 & \bfseries 0.398 \\
ViT-S In21K In1k & 0.515 & 0.362 & 0.828 & 0.700 & 0.145 & 0.510 & 0.512 & 0.301 & 0.830 & 0.441 & 0.161 & \bfseries 0.449 \\
ViT In21K In1k & 0.405 & 0.245 & 0.759 & 0.319 & 0.058 & \bfseries 0.357 & 0.425 & 0.215 & 0.784 & 0.393 & 0.061 & 0.376 \\
ViT-L In21K In1k & 0.322 & 0.105 & 0.607 & 0.205 & 0.028 & \bfseries 0.253 & 0.272 & 0.120 & 0.625 & 0.299 & 0.028 & 0.269 \\
DeiT3 & 0.505 & 0.270 & 0.829 & 0.379 & 0.183 & 0.433 & 0.437 & 0.256 & 0.824 & 0.353 & 0.171 & \bfseries 0.408 \\
DeiT3 FB In22k In1k & 0.480 & 0.236 & 0.841 & 0.386 & 0.092 & 0.407 & 0.457 & 0.243 & 0.825 & 0.395 & 0.104 & \bfseries 0.405 \\
DeiT3 In21k In1k & 0.432 & 0.201 & 0.780 & 0.388 & 0.081 & 0.376 & 0.407 & 0.206 & 0.769 & 0.360 & 0.086 & \bfseries 0.366 \\
DeiT3-L In22k In1k & 0.402 & 0.187 & 0.744 & 0.443 & 0.054 & \bfseries 0.366 & 0.405 & 0.216 & 0.792 & 0.402 & 0.063 & 0.376 \\
EVA02 & 0.691 & 0.340 & 0.837 & 0.422 & 0.379 & 0.534 & 0.515 & 0.252 & 0.872 & 0.444 & 0.116 & \bfseries 0.440 \\
EVA02 In1k & 0.418 & 0.186 & 0.800 & 0.323 & 0.153 & \bfseries 0.376 & 0.431 & 0.226 & 0.876 & 0.363 & 0.118 & 0.403 \\
EVA02 In21k & 0.638 & 0.527 & 0.805 & 0.545 &  0.308 & 0.565 & 0.393 & 0.225 & 0.751 & 0.362 & 0.106 & \bfseries 0.367 \\
EVA02-S In22k In1k & 0.526 & 0.260 & 0.800 & 0.374 & 0.152 & \bfseries 0.422 & 0.517 & 0.278 & 0.860 & 0.388 & 0.173 & 0.443 \\
EVA02 In21k In1k & 0.424 & 0.262 & 0.763 & 0.412 & 0.179 & 0.408 & 0.425 & 0.262 & 0.806 & 0.363 & 0.159 & \bfseries 0.403 \\
EVA02-L In22k In1k & 0.356 & 0.214 & 0.651 & 0.353 & 0.276 & 0.370 & 0.321 & 0.190 & 0.703 & 0.300 & 0.118 & \bfseries 0.326 \\
ViT CLIP In1k & 0.476 & 0.221 & 0.792 & 0.379 & 0.144 & \bfseries 0.402 & 0.446 & 0.222 & 0.794 & 0.387 & 0.164 & \bfseries 0.403 \\
ViT CLIP In12k In1k & 0.379 & 0.190 & 0.659 & 0.357 & 0.091 & 0.335 & 0.373 & 0.157 & 0.693 & 0.331 & 0.071 & \bfseries 0.325 \\
ViT-L CLIP In12k In1k & 0.320 & 0.167 & 0.583 & 0.375 & 0.038 & 0.297 & 0.299 & 0.159 & 0.631 & 0.326 & 0.047 & \bfseries 0.292 \\
\bottomrule
\end{tabular}

}
\label{tab:cross_model_md_performance}
\end{table}

\clearpage

\subsection{Detailed OOD performance}
\label{app:detailed-ood-performance}

Table~\ref{tab:mean_cross_model_performance} reports FPR@95 ($\downarrow$) for each (model, detector) pair together with the corresponding validation accuracy.
Across a wide range of backbones, the radially scaled (RS) variants provide the most reliable gains within the Mahalanobis families: whenever RS-MD or RS-RMD is compared against its direct baselines (MD++/MD or RMD++/RMD), it frequently attains a lower FPR@95, indicating that radially scaled normalization is a strong and generally beneficial modification.
Although a few models still achieve their absolute best score with non-Mahalanobis detectors (e.g., MLS or VIM in isolated cases), the RS variants remain consistently competitive and, in aggregate, deliver the lowest mean FPR@95 across models. This pattern suggests that the RS mechanism improves robustness across heterogeneous pretrained representations, whereas accuracy alone is not predictive of OOD performance; high-accuracy models can still exhibit higher FPR@95 than lower-accuracy ones, motivating direct evaluation of OOD detection rather than using accuracy as a proxy.

\begin{table}[ht]
\caption{
\textbf{FPR@95 ($\downarrow$) and validation accuracy (Acc, \%) across models and detectors.} Bold denotes the lowest FPR@95 \emph{within each model} across all detectors. Light green highlights cases where RS-MD outperforms both MD++ and MD, and where RS-RMD outperforms both RMD++ and RMD; in these cases, the corresponding baselines are grayed out. The Avg row reports the mean FPR@95 across models for each detector.
}
\label{tab:mean_cross_model_performance}
\adjustbox{max width=0.99\textwidth}{
\begin{tabular}{l@{\hspace{10pt}}c@{\hspace{20pt}}cccccccccc}
\toprule
 Model & Acc & MSP & MLS & KNN & VIM & RS-MD & MD++ & MD & RS-RMD & RMD++ & RMD \\
\midrule
BEiTV2 FT In1k & 85.5 & 52.2 & 50.7 & 42.6 & 39.3 & {\cellcolor[HTML]{D9F2D9}} 37.2 & \color{gray} 37.6 & \color{gray} 40.2 & \bfseries {\cellcolor[HTML]{D9F2D9}} 37.1 & \color{gray} 37.3 & \color{gray} 39.1 \\
BEiTV2 FT In21k & 85.1 & 38.0 & \bfseries 25.8 & 35.1 & 29.5 & 29.8 & 29.8 & 43.6 & 32.5 & 32.5 & 33.1 \\
DINOV2 & 83.0 & 44.9 & 32.8 & 40.2 & \bfseries 28.9 & 33.8 & 34.5 & 33.8 & 41.4 & 41.2 & 41.4 \\
MAE FT In1k & 83.5 & 54.2 & 55.7 & 44.4 & 40.6 & {\cellcolor[HTML]{D9F2D9}} 39.9 & \color{gray} 40.3 & \color{gray} 43.5 & \bfseries {\cellcolor[HTML]{D9F2D9}} 38.9 & \color{gray} 39.3 & \color{gray} 41.7 \\
ViT & 77.1 & 56.5 & 50.4 & 50.0 & 53.0 & 45.5 & 45.4 & 45.7 & 44.8 & \bfseries 44.6 & 44.9 \\
ViT-S In21K In1k & 75.8 & 57.2 & 44.4 & 53.8 & \bfseries 41.2 & 42.6 & 41.6 & 51.0 & 45.4 & 45.2 & 44.9 \\
ViT In21K In1k & 78.5 & 53.7 & 40.7 & 47.7 & 36.1 & 35.8 & 38.7 & \bfseries 35.7 & {\cellcolor[HTML]{D9F2D9}} 37.5 & \color{gray} 37.6 & \color{gray} 37.6 \\
ViT-L In21K In1k & 83.6 & 44.8 & 29.8 & 34.3 & \bfseries 25.0 & {\cellcolor[HTML]{D9F2D9}} 25.3 & \color{gray} 28.2 & \color{gray} 25.3 & 26.9 & 26.9 & 26.9 \\
ViT CLIP In1k & 84.7 & 55.2 & 65.3 & 41.1 & 41.7 & \bfseries {\cellcolor[HTML]{D9F2D9}} 37.6 & \color{gray} 38.2 & \color{gray} 40.2 & {\cellcolor[HTML]{D9F2D9}} 38.2 & \color{gray} 38.6 & \color{gray} 40.3 \\
ViT CLIP In12k In1k & 85.4 & 49.0 & 51.9 & 32.5 & 30.0 & \bfseries {\cellcolor[HTML]{D9F2D9}} 26.4 & \color{gray} 27.8 & \color{gray} 33.5 & {\cellcolor[HTML]{D9F2D9}} 30.7 & \color{gray} 30.9 & \color{gray} 32.5 \\
ViT-L CLIP In12k In1k & 86.1 & 45.0 & 43.6 & 30.1 & 28.0 & \bfseries {\cellcolor[HTML]{D9F2D9}} 26.7 & \color{gray} 27.1 & \color{gray} 29.7 & {\cellcolor[HTML]{D9F2D9}} 27.0 & \color{gray} 27.7 & \color{gray} 29.3 \\
EVA02 & 82.0 & 49.1 & 39.5 & 54.7 & \bfseries 38.8 & {\cellcolor[HTML]{D9F2D9}} 40.6 & \color{gray} 44.6 & \color{gray} 53.4 & {\cellcolor[HTML]{D9F2D9}} 44.0 & \color{gray} 44.0 & \color{gray} 44.0 \\
EVA02 FT In1k & 84.2 & 53.2 & 55.3 & 40.6 & 43.9 & \bfseries {\cellcolor[HTML]{D9F2D9}} 37.2 & \color{gray} 37.4 & \color{gray} 37.6 & {\cellcolor[HTML]{D9F2D9}} 39.6 & \color{gray} 39.8 & \color{gray} 40.3 \\
EVA02 FT In21k & 80.2 & 44.6 & 34.3 & 55.0 & \bfseries 32.9 & 50.6 & 50.6 & 56.5 & 37.7 & 37.7 & 36.7 \\
EVA02-S FT In22k In1k & 82.2 & 59.2 & 64.8 & 44.2 & 44.8 & \bfseries {\cellcolor[HTML]{D9F2D9}} 40.8 & \color{gray} 41.0 & \color{gray} 42.2 & {\cellcolor[HTML]{D9F2D9}} 43.3 & \color{gray} 43.6 & \color{gray} 44.3 \\
EVA02 FT In21k In1k & 82.2 & 53.0 & 58.9 & 42.3 & \bfseries 37.1 & 39.5 & 38.2 & 40.8 & {\cellcolor[HTML]{D9F2D9}} 38.6 & \color{gray} 39.1 & \color{gray} 40.3 \\
EVA02-L FT In22k In1k & 84.8 & 43.8 & 43.0 & 38.4 & 36.9 & 40.2 & 37.8 & 37.0 & \bfseries {\cellcolor[HTML]{D9F2D9}} 31.6 & \color{gray} 31.8 & \color{gray} 32.6 \\
DeiT3 & 83.5 & 55.0 & 59.2 & 47.5 & 47.2 & 43.2 & 43.0 & 43.3 & \bfseries {\cellcolor[HTML]{D9F2D9}} 39.4 & \color{gray} 39.9 & \color{gray} 40.8 \\
DeiT3 In21k In1k & 85.0 & 56.7 & 64.3 & 37.0 & 37.5 & {\cellcolor[HTML]{D9F2D9}} 35.5 & \color{gray} 35.6 & \color{gray} 37.6 & \bfseries {\cellcolor[HTML]{D9F2D9}} 34.5 & \color{gray} 35.1 & \color{gray} 36.6 \\
DeiT3-L In22k In1k & 85.7 & 58.1 & 65.9 & 35.9 & 39.7 & \bfseries {\cellcolor[HTML]{D9F2D9}} 33.4 & \color{gray} 34.2 & \color{gray} 36.6 & {\cellcolor[HTML]{D9F2D9}} 35.5 & \color{gray} 35.9 & \color{gray} 37.6 \\
DeiT3 FB In22k In1k & 83.8 & 60.9 & 64.9 & 41.1 & 39.8 & 40.2 & \bfseries 38.8 & 40.7 & {\cellcolor[HTML]{D9F2D9}} 39.2 & \color{gray} 39.2 & \color{gray} 40.5 \\
\hline
Avg &  & 51.6 &  49.6 & 42.3 &  37.7 & \bfseries 37.2 & 37.6 & \bfseries 40.4 & \bfseries 37.3 &  37.5 & 38.3 \\
\bottomrule
\end{tabular}

}   
\end{table}

\clearpage

\begin{table}[!ht]
\centering
\caption{
\textbf{FPR on NINCO across model families for Mahalanobis variants} (lower is better).
\textbf{MD*} uses the empirically optimal $\beta$; $\hat{\textbf{MD}}$ uses the regression-predicted $\hat{\beta}$; \textbf{MD} (standard) fixes $\beta=0$; and \textbf{MD++} (Mahalanobis++) fixes $\beta=1$.
}
\begin{tabular}{l@{\hspace{10pt}}c@{\hspace{20pt}}cccccccccc}
\toprule
 Model & Acc & MSP & MLS & KNN & VIM & RS-MD & MD++ & MD & RS-RMD & RMD++ & RMD \\
\midrule
BEiTV2 FT In1k & 85.5 & 57.0 & 63.9 & 56.6 & 55.1 & {\cellcolor[HTML]{D9F2D9}} 46.3 & \color{gray} 47.0 & \color{gray} 50.6 & \bfseries {\cellcolor[HTML]{D9F2D9}} 43.9 & \color{gray} 44.3 & \color{gray} 45.6 \\
BEiTV2 FT In21k & 85.1 & 40.8 & \bfseries 28.7 & 44.3 & 33.0 & 36.4 & 36.4 & 47.5 & 33.1 & 33.1 & 33.9 \\
DINOV2 & 83.0 & 48.3 & 37.2 & 54.7 & \bfseries 35.2 & 42.4 & 44.5 & 42.4 & 57.7 & 57.4 & 57.7 \\
MAE FT In1k & 83.5 & 56.4 & 67.0 & 51.9 & 51.8 & {\cellcolor[HTML]{D9F2D9}} 48.5 & \color{gray} 48.8 & \color{gray} 51.6 & \bfseries {\cellcolor[HTML]{D9F2D9}} 44.6 & \color{gray} 45.0 & \color{gray} 46.3 \\
ViT & 77.1 & 61.8 & 63.5 & 58.7 & 71.4 & {\cellcolor[HTML]{D9F2D9}} 55.6 & \color{gray} 55.6 & \color{gray} 55.7 & 51.4 & \bfseries 51.4 & 51.6 \\
ViT-S In21K In1k & 75.8 & 60.3 & 54.1 & 60.1 & 52.0 & 53.1 & \bfseries 50.5 & 51.5 & 51.3 & 51.4 & 51.2 \\
ViT In21K In1k & 78.5 & 61.1 & 49.3 & 54.6 & 46.6 & 41.7 & 48.1 & \bfseries 40.5 & {\cellcolor[HTML]{D9F2D9}} 42.2 & \color{gray} 42.5 & \color{gray} 42.5 \\
ViT-L In21K In1k & 83.6 & 45.9 & 34.0 & 41.5 & 30.9 & {\cellcolor[HTML]{D9F2D9}} 31.0 & \color{gray} 35.8 & \color{gray} 32.2 & \bfseries {\cellcolor[HTML]{D9F2D9}} 26.9 & \color{gray} 27.7 & \color{gray} 27.2 \\
ViT CLIP In1k & 84.7 & 59.2 & 79.4 & 46.6 & 54.9 & {\cellcolor[HTML]{D9F2D9}} 44.8 & \color{gray} 45.5 & \color{gray} 47.6 & \bfseries {\cellcolor[HTML]{D9F2D9}} 42.8 & \color{gray} 43.3 & \color{gray} 44.6 \\
ViT CLIP In12k In1k & 85.4 & 49.2 & 68.5 & 39.1 & 35.4 & \bfseries {\cellcolor[HTML]{D9F2D9}} 31.7 & \color{gray} 33.6 & \color{gray} 37.9 & {\cellcolor[HTML]{D9F2D9}} 34.5 & \color{gray} 35.2 & \color{gray} 37.3 \\
ViT-L CLIP In12k In1k & 86.1 & 45.4 & 59.1 & 33.7 & 32.9 & 33.5 & 31.0 & 32.0 & \bfseries {\cellcolor[HTML]{D9F2D9}} 28.4 & \color{gray} 29.0 & \color{gray} 29.9 \\
EVA02 & 82.0 & 54.7 & \bfseries 50.5 & 63.2 & 58.7 & {\cellcolor[HTML]{D9F2D9}} 59.0 & \color{gray} 63.0 & \color{gray} 69.1 & 52.8 & 52.3 & 51.5 \\
EVA02 FT In1k & 84.2 & 58.8 & 75.5 & 46.8 & 71.0 & \bfseries {\cellcolor[HTML]{D9F2D9}} 41.0 & \color{gray} 41.3 & \color{gray} 41.8 & {\cellcolor[HTML]{D9F2D9}} 42.2 & \color{gray} 42.5 & \color{gray} 43.1 \\
EVA02 FT In21k & 80.2 & 47.1 & 45.8 & 56.5 & 40.7 & 56.3 & 56.3 & 63.8 & 40.0 & 40.0 & \bfseries 39.3 \\
EVA02-S FT In22k In1k & 82.2 & 67.2 & 82.5 & 54.5 & 69.4 & \bfseries {\cellcolor[HTML]{D9F2D9}} 49.9 & \color{gray} 50.4 & \color{gray} 52.6 & {\cellcolor[HTML]{D9F2D9}} 51.2 & \color{gray} 51.3 & \color{gray} 51.7 \\
EVA02 FT In21k In1k & 82.2 & 57.2 & 82.7 & 46.1 & 50.9 & 43.4 & 40.7 & 42.4 & \bfseries {\cellcolor[HTML]{D9F2D9}} 40.5 & \color{gray} 40.6 & \color{gray} 42.5 \\
EVA02-L FT In22k In1k & 84.8 & 48.0 & 68.4 & 37.0 & 59.5 & 41.7 & 38.8 & 35.6 & \bfseries {\cellcolor[HTML]{D9F2D9}} 31.5 & \color{gray} 31.6 & \color{gray} 32.1 \\
DeiT3 & 83.5 & 58.5 & 70.4 & 55.8 & 63.6 & {\cellcolor[HTML]{D9F2D9}} 49.9 & \color{gray} 50.0 & \color{gray} 50.5 & \bfseries {\cellcolor[HTML]{D9F2D9}} 43.2 & \color{gray} 43.3 & \color{gray} 43.7 \\
DeiT3 In21k In1k & 85.0 & 64.9 & 86.3 & 44.8 & 54.1 & 42.6 & 42.3 & 43.2 & \bfseries {\cellcolor[HTML]{D9F2D9}} 37.8 & \color{gray} 38.7 & \color{gray} 40.7 \\
DeiT3-L In22k In1k & 85.7 & 65.2 & 84.8 & 40.1 & 55.8 & 39.0 & 38.8 & 40.2 & \bfseries {\cellcolor[HTML]{D9F2D9}} 38.1 & \color{gray} 38.5 & \color{gray} 40.5 \\
DeiT3 FB In22k In1k & 83.8 & 63.8 & 82.0 & 47.7 & 57.8 & 47.8 & 45.5 & 48.0 & 44.5 & \bfseries 44.5 & 45.7 \\
\hline
Avg &  & 55.8 &  63.5 &  49.3 &  51.5 & \bfseries 44.5 &  45.0 &  46.5 & \bfseries 41.8 & 42.1 & 42.8 \\
\bottomrule
\end{tabular}

\label{tab:app_ninco_conformal_md}
\end{table}

\clearpage

\section{Geometry of Eigvenvalues}
\label{app:eigenvalue_plots_geometry}

\begin{figure}[h!]
    \includegraphics[width=\linewidth]{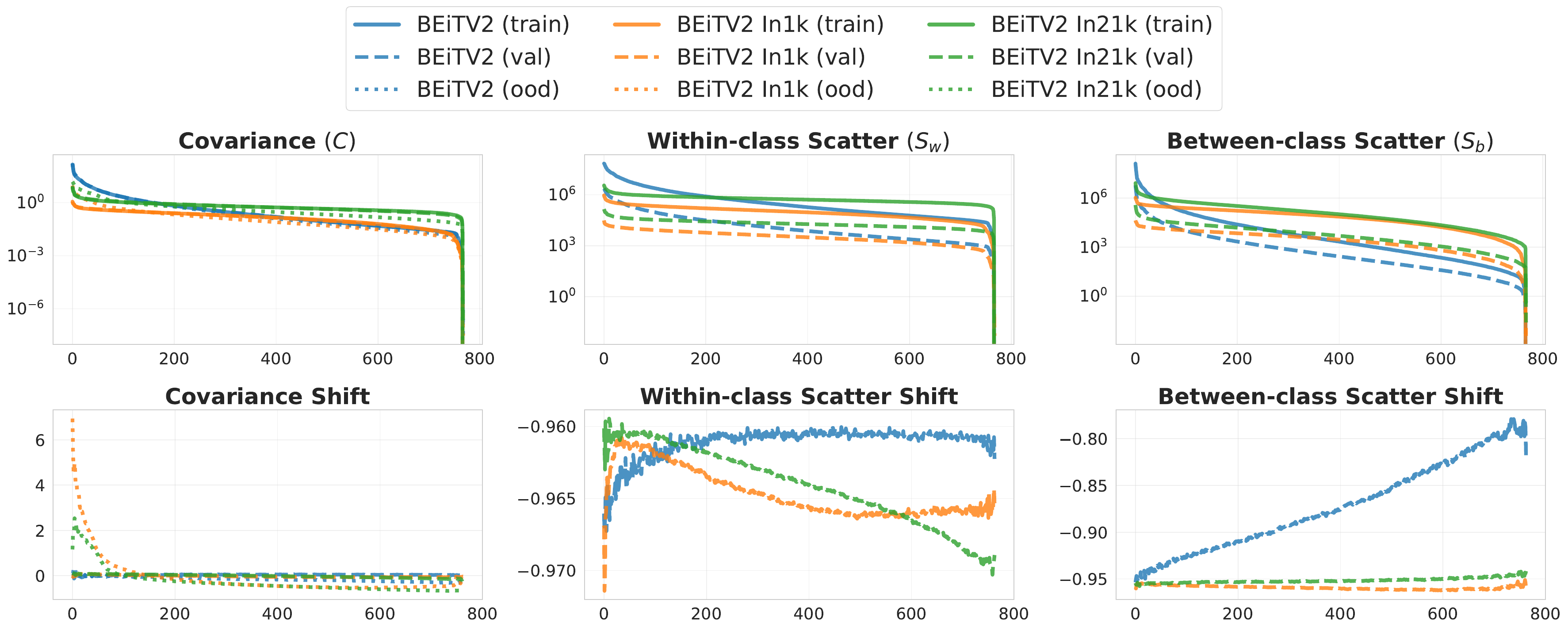}
    \caption{
    BEiTV2 eigenspectra and their respective shifts: top—eigenvalues of covariance C, within-class $S_w$, and between-class $S_b$ across train (solid), val (dashed), and OOD (dotted); bottom—corresponding OOD-induced eigenvalue shifts relative to train.
    }
    \label{fig:beit_eigvals_appendix}
\end{figure}

\begin{figure}[h!]
    \includegraphics[width=\linewidth]{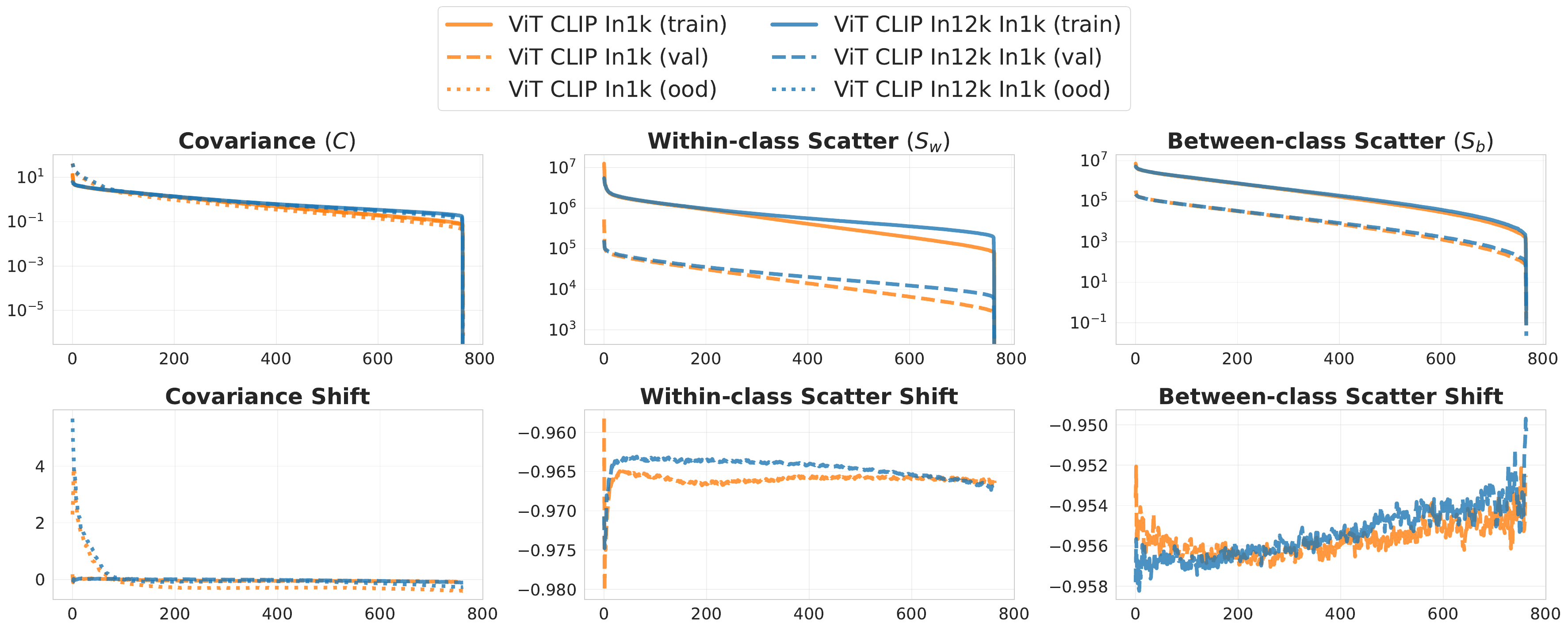}
    \caption{
    CLIP eigenspectra and their respective shifts: top—eigenvalues of covariance C, within-class $S_w$, and between-class $S_b$ across train (solid), val (dashed), and OOD (dotted); bottom—corresponding OOD-induced eigenvalue shifts relative to train.
    }
    \label{fig:clip_eigvals_appendix}
\end{figure}

\begin{figure}[h!]
    \includegraphics[width=\linewidth]{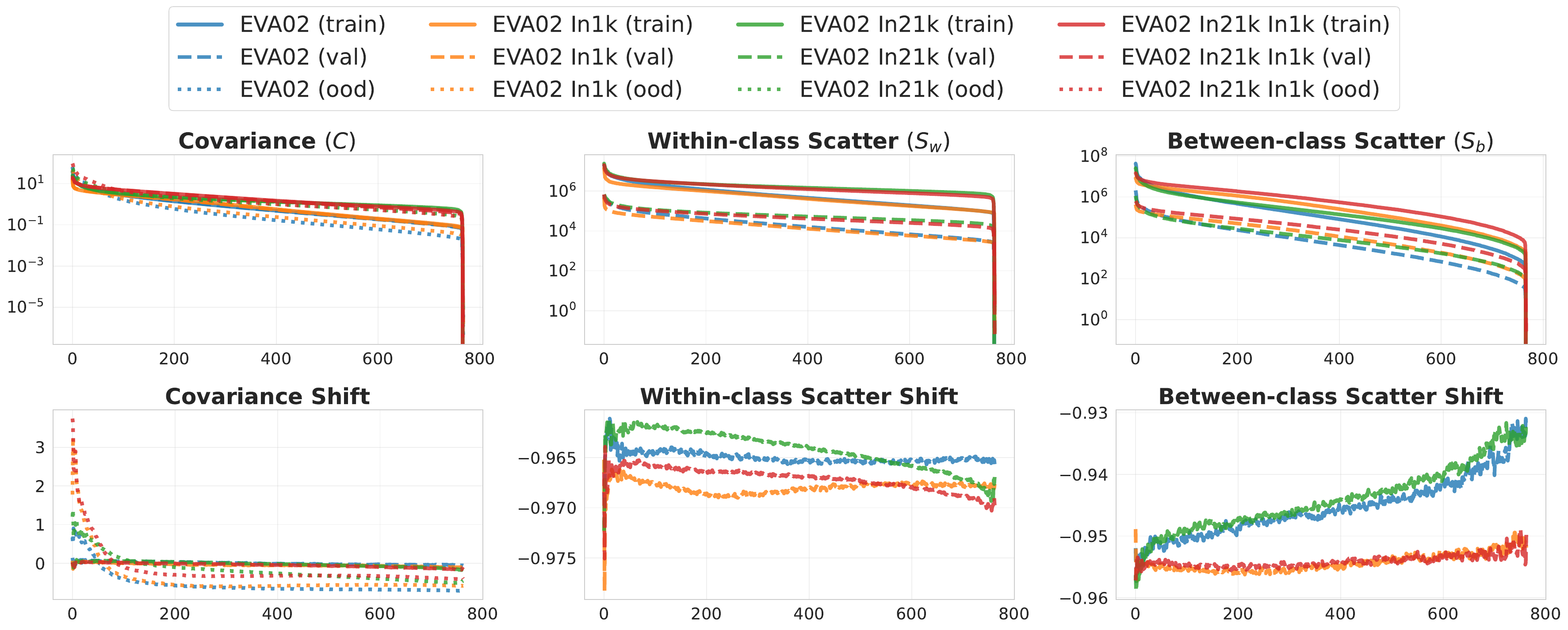}
    \caption{
        EVA02 eigenspectra and their respective shifts: top—eigenvalues of covariance C, within-class $S_w$, and between-class $S_b$ across train (solid), val (dashed), and OOD (dotted); bottom—corresponding OOD-induced eigenvalue shifts relative to train.
    }
    \label{fig:eva_eigvals_appendix}
\end{figure}

\begin{figure}[h!]
    \includegraphics[width=\linewidth]{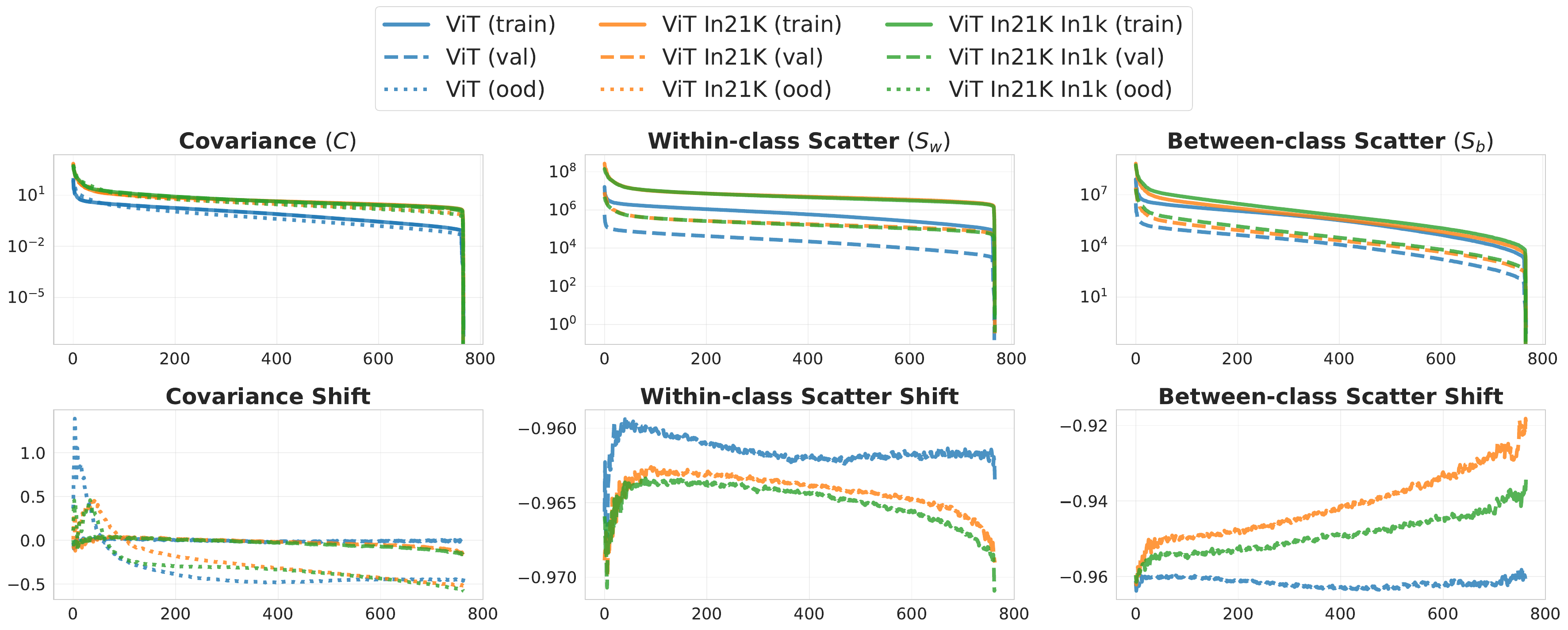}
    \caption{
        ViT eigenspectra and their respective shifts: top—eigenvalues of covariance C, within-class $S_w$, and between-class $S_b$ across train (solid), val (dashed), and OOD (dotted); bottom—corresponding OOD-induced eigenvalue shifts relative to train.
    }
    \label{fig:vit_eigvals_appendix}
\end{figure}

\begin{figure}[h!]
    \centering
    \includegraphics[width=0.6\linewidth]{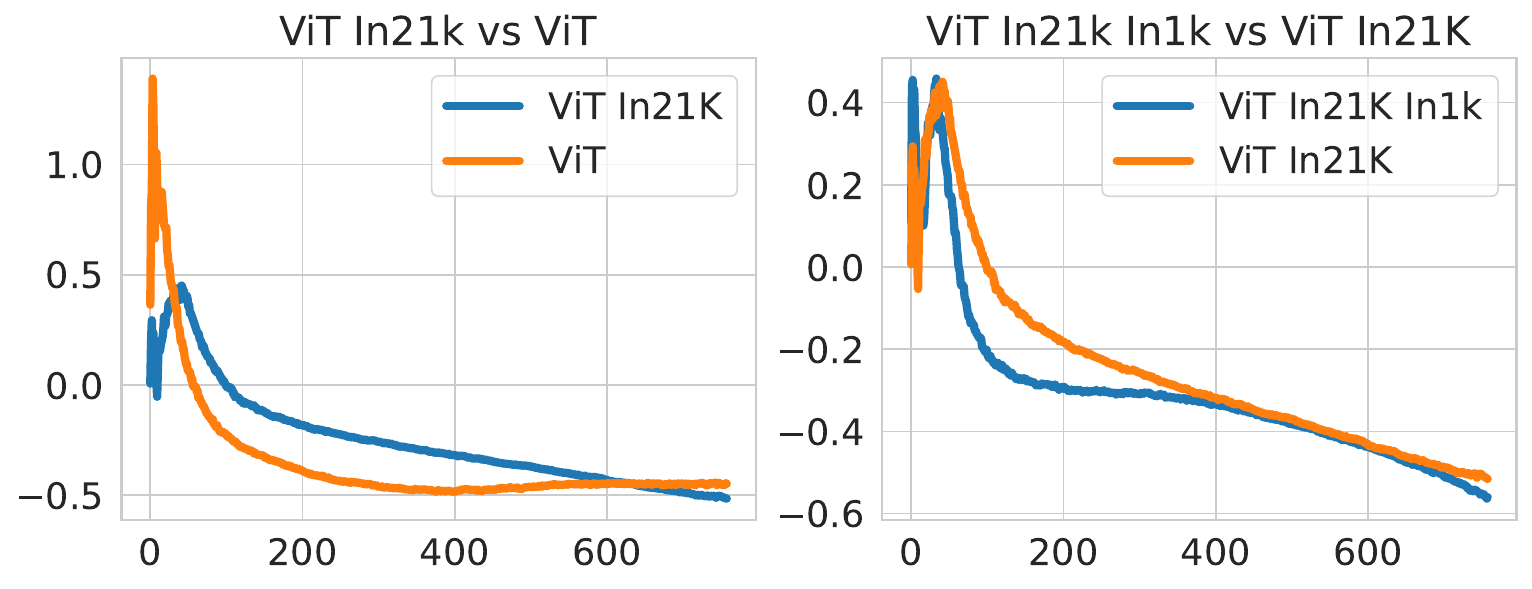}
    \caption{Eigenspectrum of covariance shift between train and OOD data (NINCO) for ViT variants: left—ViT In21K vs ViT; right—ViT In21K In1k vs ViT In21K.
    }
    \label{fig:eigvals_shift_vit_appendix}
\end{figure}

\begin{figure}[h!]
    \centering
    \includegraphics[height=0.9\textheight,keepaspectratio]{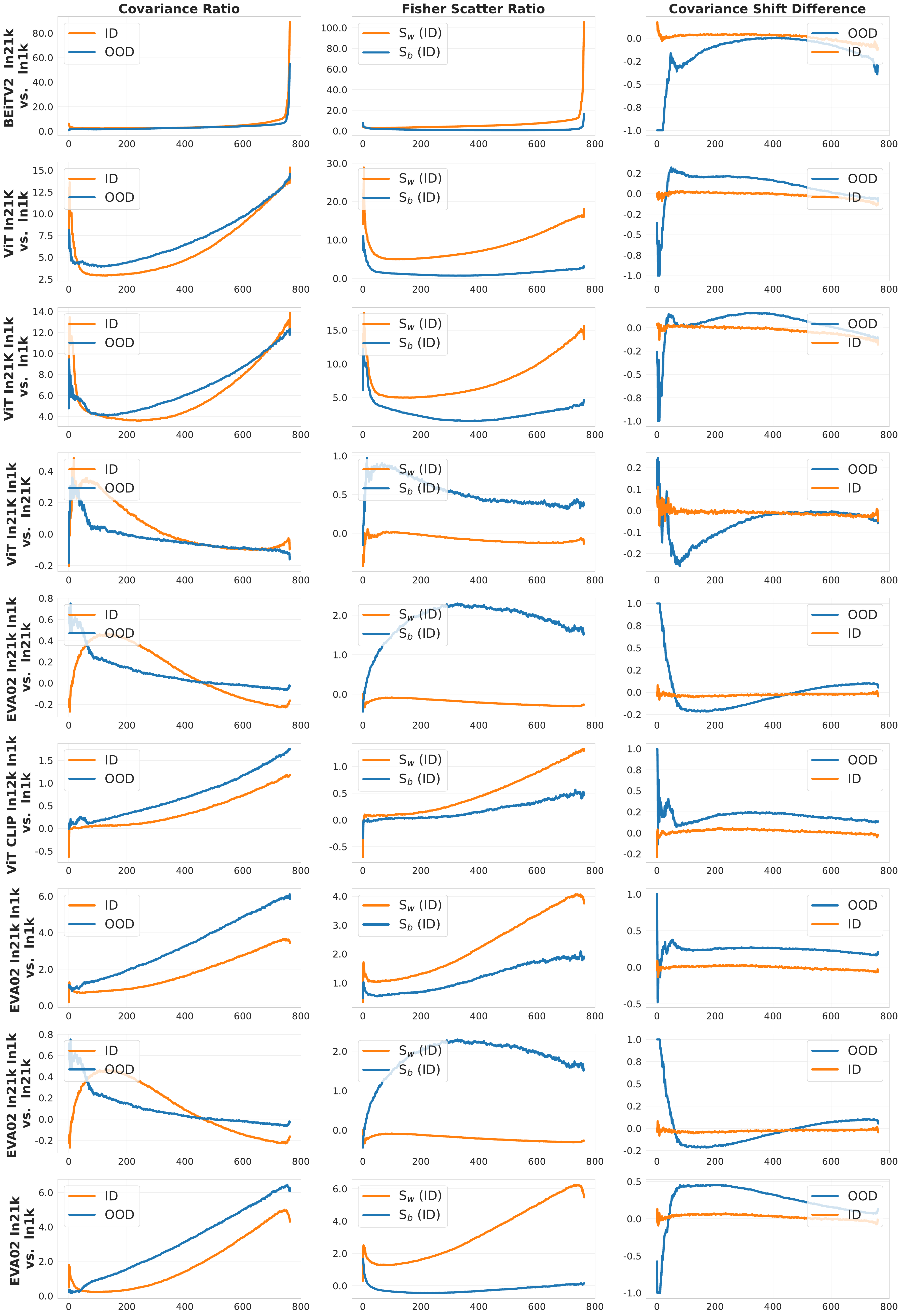}
    \caption{Eigenspectrum differences by model pair (BEiTV2, ViT, EVA02/CLIP): for each pair, we plot ID vs OOD covariance $C$, ID within‑class $S_w$ and between‑class $S_b$, and covariance‑shift curves (OOD and ID), showing relative eigenvalue changes between the first and second model.
    }
    \label{fig:eigvals_diff_by_model_appendix}
\end{figure}

\clearpage

\section{Pearson correlations}
\label{app:pearson-correlation}

\begin{figure}[h!]
    \centering
    \includegraphics[width=0.8\linewidth]{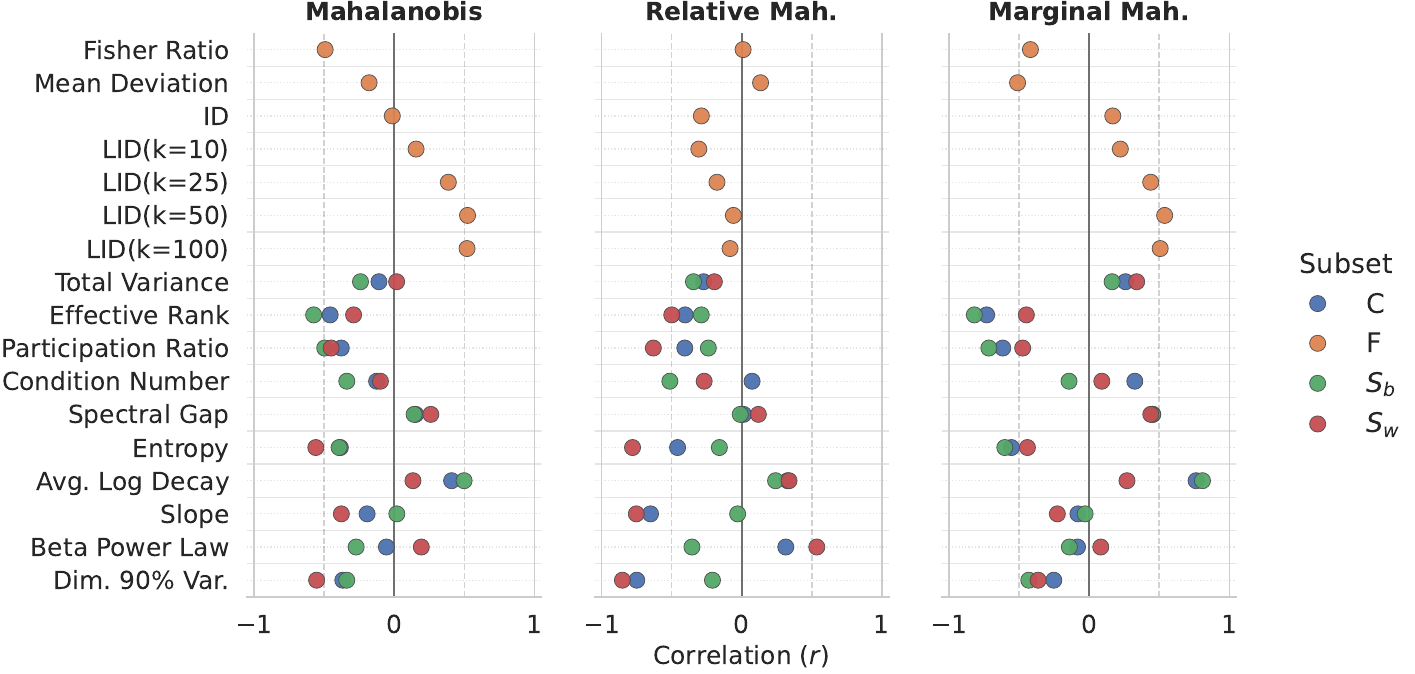}
    \caption{Pearson correlations between representation metrics and OOD performance across Mahalanobis variants. The three Mahalanobis-based detectors exploit different geometric cues, leading to distinct correlation patterns consistent with the Spearman results in Figure~\ref{fig:correlation_heatmap_spearman}.}
    \label{fig:correlation_heatmap_pearson}
    \vspace{-1.0em}
\end{figure}

We replicate the correlation analysis from the main paper using Pearson correlations instead of Spearman correlations (Figure \ref{fig:correlation_heatmap_pearson}). The trends remain consistent: manifold-geometry and eigenvalue-based metrics show similar relationships with OOD performance across the three Mahalanobis variants, confirming that the observed patterns are not sensitive to the choice of correlation metric.

\clearpage

\section{Full model names}
\label{app:full_model_names}

\begin{table}[!ht]
\centering
\caption{Mapping of model names to checkpoints and sources.}
\adjustbox{max width=0.98\textwidth}{
\begin{tabular}{lll}
\hline
\textbf{Model Name} & \textbf{Checkpoint (Version)} & \textbf{Source} \\
\hline
BEiTV2 In1k & beitv2\_base\_patch16\_224.in1k\_ft\_in1k & timm / huggingface \\
BEiTV2 In21k & beitv2\_base\_patch16\_224.in1k\_ft\_in22k & timm / huggingface \\
DINOV2 & vit\_base\_patch14\_dinov2.lvd142m & timm / huggingface \\
DINOV3 & dinov3-vitb16-pretrain-lvd1689m & facebook / huggingface \\
MAE In1k & mae\_finetuned\_vit\_base & github.com/facebookresearch/mae \\
ViT & vit\_base\_patch16\_224.augreg\_in1k & timm / huggingface \\
ViT In21K & vit\_base\_patch16\_224.augreg\_in21k & timm / huggingface \\
ViT In21K In1k & vit\_base\_patch16\_224.augreg\_in21k\_ft\_in1k & timm / huggingface \\
ViT-S In21K In1k & vit\_small\_patch16\_224.augreg\_in21k\_ft\_in1k & timm / huggingface \\
ViT-L In21K In1k & vit\_large\_patch16\_224.augreg\_in21k\_ft\_in1k & timm / huggingface \\
ViT CLIP In1k & vit\_base\_patch16\_clip\_224.laion2b\_ft\_in1k & timm / huggingface \\
ViT CLIP In12k In1k & vit\_base\_patch16\_clip\_224.laion2b\_ft\_in12k\_in1k & timm / huggingface \\
ViT-L CLIP In12k In1k & vit\_large\_patch14\_clip\_336.laion2b\_ft\_in12k\_in1k & timm / huggingface \\
EVA02 & eva02\_base\_patch14\_224.mim\_in22k & timm / huggingface \\
EVA02 In1k & eva02\_base\_patch14\_448.mim\_in22k\_ft\_in1k & timm / huggingface \\
EVA02 In21k & eva02\_base\_patch14\_448.mim\_in22k\_ft\_in22k & timm / huggingface \\
EVA02 In21k In1k & eva02\_base\_patch14\_448.mim\_in22k\_ft\_in22k\_in1k & timm / huggingface \\
EVA02-L In22k In1k & eva02\_large\_patch14\_448.mim\_m38m\_ft\_in22k\_in1k & timm / huggingface \\
EVA02-S In22k In1k & eva02\_small\_patch14\_336.mim\_in22k\_ft\_in1k & timm / huggingface \\
DeiT3 & deit3\_base\_patch16\_224 & timm / huggingface \\
DeiT3 In21k In1k & deit3\_base\_patch16\_224\_in21ft1k & timm / huggingface \\
DeiT3 FB In22k In1k & deit3\_base\_patch16\_384.fb\_in22k\_ft\_in1k & timm / huggingface \\
DeiT3-L In22k In1k & deit3\_large\_patch16\_384.fb\_in22k\_ft\_in1k & timm / huggingface \\
\hline
\end{tabular}

}
\label{tab:model_checkpoints}
\end{table}

\section{Use of AI Assistance}
AI assistants, such as ChatGPT, were utilized in various aspects of the research, including coding, data analysis, and writing tasks. These tools helped automate repetitive tasks, generate initial drafts, and assist in exploring potential solutions. However, all AI-generated outputs were reviewed and refined by researchers to ensure accuracy and coherence.

\end{document}